\documentclass[10pt, a4paper]{article}

\usepackage[final]{lrec2026} 
\usepackage{tcolorbox}
\usepackage{booktabs}
\usepackage{adjustbox}
\usepackage{amssymb}
\usepackage{times}
\usepackage{latexsym}
\usepackage{url}
\usepackage{multirow}
\usepackage{booktabs}
\usepackage[table,xcdraw]{xcolor}
\usepackage{colortbl}
\usepackage{xfp} 
\usepackage{graphicx}
\usepackage{subcaption}
\usepackage{float} 
\definecolor{mygreen}{HTML}{659157}
\definecolor{myred}{HTML}{987284}
\usepackage[T1]{fontenc}

\usepackage[utf8]{inputenc}

\usepackage{microtype}
\usepackage{inconsolata}

\usepackage{graphicx}

\title{    \includegraphics[height=1em]{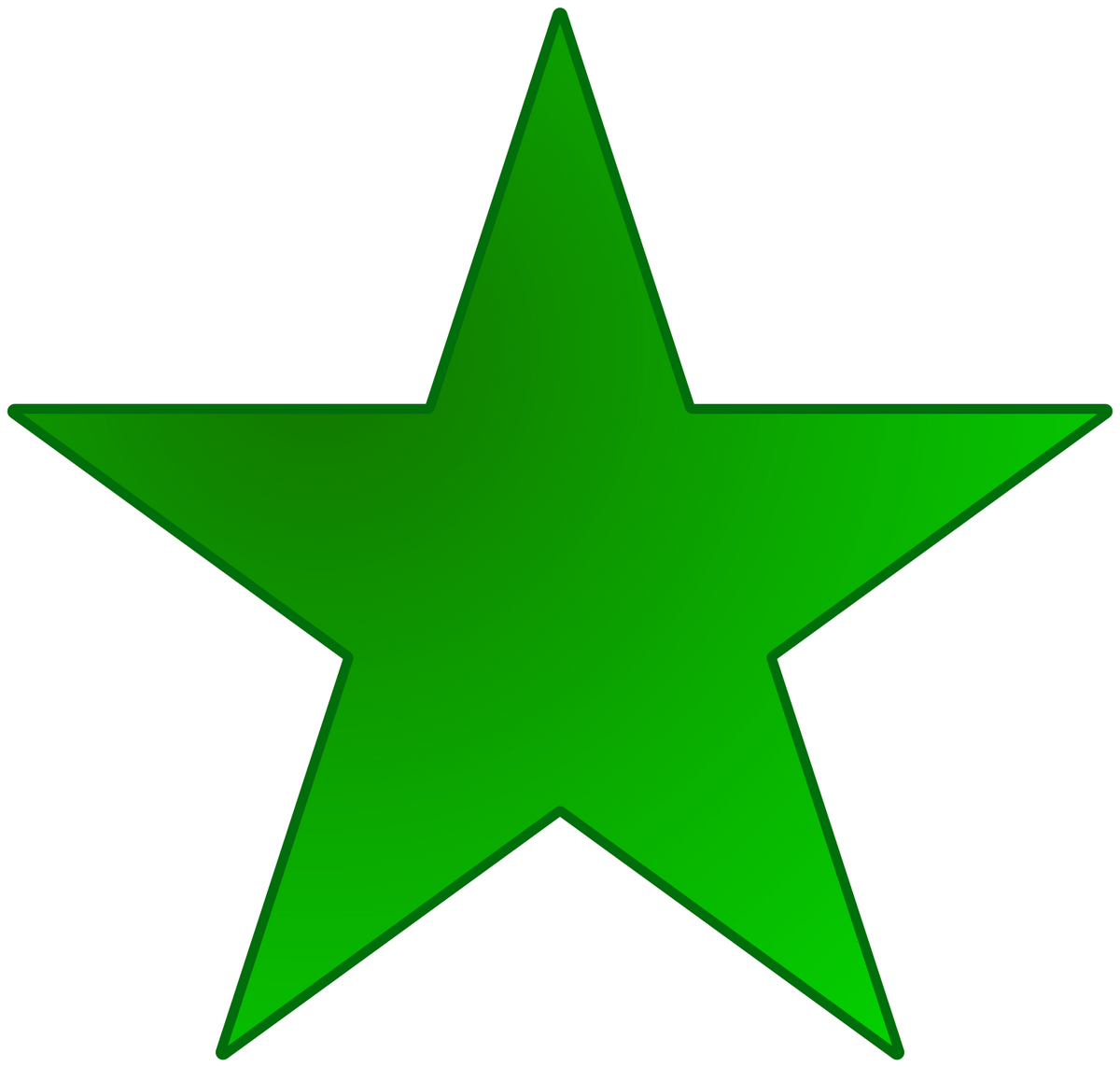}Open Machine Translation for Esperanto}

\name{Ona de Gibert$^{\heartsuit}$, Lluís de Gibert$^{\bigstar}$} 

\address{\\$^{\heartsuit}$University of Helsinki, Department of Digital Humanities \\
        $^{\bigstar}$Kataluna Esperanto-Asocio (KEA) \\
        $^{\bigstar}$Sennacieca Asocio Tutmonda (SAT) \\
        \texttt{ona.degibert@helsinki.fi}\\}

\abstract{
Esperanto is a widespread constructed language, known for its regular grammar and productive word formation. Besides having substantial resources available thanks to its online community, it remains relatively underexplored in the context of modern machine translation (MT) approaches.
In this work, we present the first comprehensive evaluation of open-source MT systems for Esperanto, comparing rule-based systems, encoder–decoder models, and LLMs across model sizes. We evaluate translation quality across six language directions involving English, Spanish, Catalan, and Esperanto using multiple automatic metrics as well as human evaluation.
Our results show that the NLLB family achieves the best performance in all language pairs, followed closely by our trained compact models and a fine-tuned general-purpose LLM.
Human evaluation confirms this trend, with NLLB translations preferred in approximately half of the comparisons, although noticeable errors remain.
In line with Esperanto’s tradition of openness and international collaboration, we release our code and best-performing models publicly.
\\ \newline }

\begin{document}

\maketitleabstract

\section{Introduction}
Constructed languages (\textit{conlangs}) are languages intentionally created for human communication rather than emerging through natural linguistic evolution \cite{kuhn-2014-survey}. From the perspective of language technology, conlangs occupy an unusual position: they typically attract limited commercial investment, which reduces incentives to develop dedicated tools and resources \cite{occhini2026artificial}. At the same time, many conlangs are supported by active online communities and maintain a considerable web presence, which leads to their inclusion in large-scale training corpora. Among conlangs, Esperanto is the most prominent and widely used example \citep{blanke2009causes}. 

Esperanto represents a unique case among constructed languages. It has a well-developed Wikipedia with over 380,000 articles and a large global community of second-language speakers. 
In addition, Esperanto ranks 75th in language presence in Common Crawl as of the most recent crawl (CC-MAIN-2026-04).\footnote{Data Source: \url{https://commoncrawl.github.io/cc-crawl-statistics/plots/languages}} This indicates substantial web representation relative to many other low-resource natural languages. Esperanto is also included in major large-scale pretraining corpora such as MADLAD-400 \cite{kudugunta2023madlad} and HPLT \citep{de-gibert-etal-2024-new,burchell-etal-2025-expanded,oepen2025hplt}. Consequently, modern Large Language Models (LLMs) are exposed to non-trivial amounts of Esperanto during pretraining and are capable of generating it. However, the absence of dedicated evaluation benchmarks makes it difficult to systematically assess their true proficiency. 
\textcolor{black}{While Esperanto has non-trivial available resources, we frame Esperanto as under-resourced in terms of its underrepresentation in language technology and limited targeted development of NLP resources and applications.}

\begin{figure}
    \centering
    \includegraphics[width=\linewidth]{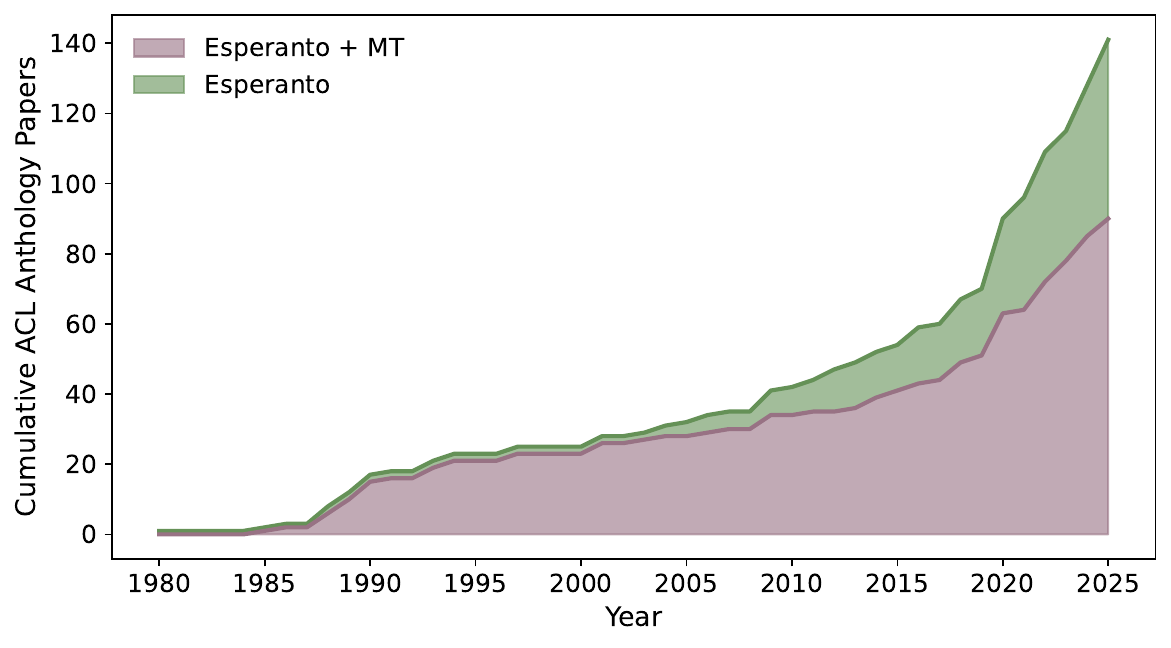}
    \caption{Cumulative number of ACL Anthology papers mentioning 'Esperanto' and 'Esperanto and Machine Translation'. Early work involving Esperanto focused primarily on MT, while more recent work covers a broader range of topics.}
    \label{fig:acl}
\end{figure}

Machine translation (MT) provides a practical and controlled evaluation framework. 
Despite the existence of substantial textual resources, there is currently no state-of-the-art, openly available MT system specifically optimized for Esperanto. 
While we are aware Esperanto is included in commercial platforms, we disregard them in this work.
We focus on evaluating and developing open MT systems for translation from Esperanto (eo) into Catalan (ca), Spanish (es) and English (en); and vice-versa. 



We present the first systematic benchmark of open-source MT systems for Esperanto translation, comparing rule-based systems, encoder–decoder models, and LLMs across model sizes ranging from 600M to 9B parameters. We find that LLMs lag behind specialized MT systems, while the NLLB family achieves the best overall performance. We further train compact Transformer models that remain competitive with substantially larger systems while being more efficient and sustainable. Finally, we conduct a human evaluation and qualitative error analysis to better understand the strengths and weaknesses of the evaluated models.


In our work, we adopt an open-science approach, as we aim to support and develop decentralized, grassroots language technology for Esperanto, reflecting its core values of linguistic equality, international communication, and community collaboration.
To support reproducibility, we release our code in a public GitHub repository\footnote{\url{https://github.com/onadegibert/EsperantoMT}} and our best models on HuggingFace\footnote{\url{https://huggingface.co/collections/Helsinki-NLP/open-machine-translation-for-esperanto}}.

\section{An Introduction to Esperanto}
\label{sec:esperanto}
Esperanto is an \textit{a posteriori} conlang  \cite{couturat1903histoire}, modeled on existing languages. It was proposed by Zamenhof in 1887 with the aim of enabling universal communication in a neutral context.
Esperanto represents the most consolidated case of a planned language, designed to be as accessible to as many speakers as possible. It is spoken in more than 100 countries \cite{poncelas2020using}. Estimates of the number of second language speakers vary considerably —from tens of thousands \cite{eberhard2026ethnologue} to several million \cite{wandel2015many}— , reflecting the methodological difficulty of obtaining precise demographic data in transnational communities.
It is also the only conlang that has developed a stable community of first language speakers, usually estimated at between 1,000 and 2,000 individuals \cite{eberhard2026ethnologue}. Despite not having official state status, Esperanto has been the subject of institutional recognition, including UNESCO resolutions since 1954 that highlight its potential as a tool for international understanding \cite{unesco1954resolution}. At the same time, its presence in digital environments and both original and translated literary production contribute to shaping an active communicative ecosystem, with intergenerational continuity and capacity for technological adaptation.

Esperanto uses the Latin alphabet with diacritics. It was designed according to principles of structural regularity and morphological transparency, which make it highly unambiguous.
From a typological perspective, it presents a highly systematic agglutinative morphology, based on the productive combination of invariable roots with a restricted and regular set of prefixes and suffixes.
This mechanism reduces morphosyntactic irregularity and favors a high degree of lexical compositionality. In terms of vocabulary, the language derives from Indo-European languages, with an approximate proportion of 80\% of Romance origin, complemented by Germanic, Slavic and Greek contributions \cite{parkvall2010european}. 
It has often been described as structurally transparent and morphologically predictable language, suitable for MT \citep{schubert2002esperanto,gobbo2015machine}. 

\section{Related Work}
\label{sec:related}

We review prior research specifically focused on Esperanto. First, to better understand the evolution of the field, we query the full ACL Anthology for the terms “Esperanto”, and “Esperanto" and "Machine Translation” using the \texttt{acl-crawl} tookit.\footnote{\url{https://github.com/Sethjsa/acl-crawl}} 
Figure~\ref{fig:acl} shows the cumulative number of papers matching our queries. Esperanto has been present in the literature since the 1980s. In the early years, most papers mentioning Esperanto focused on MT, but more recently, the topics have diverged. This trend is expected, as research on MT has declined in general and the field has increasingly shifted toward natural language understanding (NLU) tasks.

\subsection{Esperanto in NLP}
Early work on Esperanto in NLP concentrated on rule-based linguistic modeling. \citet{karlsson1990constraint} developed a Constraint Grammar framework for Esperanto, \citet{minnaja-paccagnella-2000-part} developed a Part-of-Speech tagger, and \citet{manaris2006investigating} performed a corpus-based linguistic analysis, comparing Esperanto to other European languages. \citet{bick-2016-morphological} introduced a morphological lexicon for Esperanto and later a syntactic treebank \citep{bick-2020-syntax}.
Beyond core NLP tasks, \citet{FIEDLER2018166} investigated code-switching phenomena between Esperanto and English. 
More recently, \citet{oya-2025-ud} introduced Universal Dependencies annotations for Esperanto, while \citet{bick-2025-annotated} released a learner corpus with error annotations and Constraint Grammar tags. This shows that the development of core NLP resources for Esperanto is still an active area of research.

\subsection{Esperanto in MT}
Esperanto attracted considerable attention in the early development of MT, particularly during the rule-based and statistical MT eras. Due to its regular grammar and its lexicon derived from multiple European languages, several studies proposed Esperanto as a potential interlingua for multilingual MT systems \cite{witkam-1984-distributed,neijt1986esperanto,franco-sabaris-etal-2001-multilingual,boddington2004evaluation}. However, these works were largely conceptual and exploratory rather than large-scale implemented systems.

Regarding Rule-Based MT \cite{hutchins1992introduction}, Apertium \cite{apertium}, a free and open-source toolkit for developing rule-based MT systems, has included Esperanto translation pairs since its early releases. Additionally, \citet{bick2011wikitrans} developed a rule-based system that translated portions of the English Wikipedia into Esperanto.
During the statistical MT period, Esperus \citep{orlova2015esperus} was developed as a Russian--Esperanto system using the MOSES toolkit \cite{koehn2007moses}. Other systems from that period include Esperantilo (2008), Lingvohelpilo (2009), and Lingvoilo (2015) \cite{burghelea2019not}. These systems demonstrated practical interest in Esperanto MT but remained relatively small-scale.
In recent years, dedicated research on Esperanto in MT has significantly diminished. One notable exception is \citet{poncelas2020using}, who explored tokenization strategies for literary translation between Esperanto and English.
More recently, Esperanto is included in the No Language Left Behind (NLLB) initiative \cite{costa2022no, nllb2024scaling} a highly multilingual model family, capable of translating among more than 200 languages. As a result, Esperanto is also included in the FLORES+ benchmark \cite{goyal2022flores}, which provides a standardized evaluation test set for translation between Esperanto and 200 other languages. This represents an important step forward for Esperanto in multilingual MT evaluation.
Beyond this, however, the past decade has seen limited focused research on Esperanto translation within modern neural MT paradigms. 


\section{Experimental Setup}
\label{sec:Eexperiments}
\begin{table*}[t]
\centering
\small
\begin{tabular}{lrrrrrrr}
\toprule
 & \multicolumn{3}{c}{Training Data} & \multicolumn{2}{c}{Used for Training} & \multicolumn{2}{c}{FLORES+} \\
\cmidrule(lr){2-4} \cmidrule(lr){5-6} \cmidrule(lr){7-8}
Language Pair & Raw & Filtered & \% Removed & Marian & LLM FT & Dev & Test \\
\midrule
en–eo & 70.34M & 45.39M & 35.5\% & 5.0M & 100k & 997 & 1012 \\
es–eo & 6.21M & 4.68M & 24.7\% & 4.7M & 100k & 997 & 1012 \\
ca–eo & 1.17M & 673.93k & 42.7\% & 673k & 100k & 997 & 1012 \\
\midrule
Total & 77.73M & 50.74M & &  10.35M & 300k & 2991 \\
\bottomrule
\end{tabular}
\caption{Parallel data statistics before and after filtering, and final training sizes per model family.}
\label{tab:data}

\end{table*}

We study MT models for Esperanto to translate both from and into English, Catalan, and Spanish. 
We are interested in these languages because Esperanto has a strong presence in the Iberian Peninsula, as demonstrated by the development efforts on open-source rule-based MT systems for these language pairs \cite{apertium} and active groups scattered around Catalonia, the Basque Country, Valencia and Andalusia. 

Our experimental framework is divided into two parts. We first introduce our benchmarking setup, where we evaluate existing systems out-of-the-box. We then describe our MT development efforts, where we train encoder-decoder models and fine-tune an LLM. Finally, we present the metrics employed.

\subsection{Benchmarking}

To assess the current state of Esperanto MT, we evaluate several publicly available systems without any additional fine-tuning.
This allows us to establish a realistic performance baseline and measure how well Esperanto is currently supported in multilingual MT systems.

\paragraph{Models} We evaluate models representing different architectures and modeling paradigms:

\begin{itemize}

    \item \textbf{Apertium} \cite{apertium}: 
    A rule-based MT system relying on manually crafted linguistic rules and bilingual dictionaries. Apertium is computationally efficient and lightweight. However, it only supports four of the six translation directions considered in this study.

    \item \textbf{NLLB family} \cite{costa2022no,nllb2024scaling}: 
    We evaluate four models from the \href{https://huggingface.co/facebook/models?search=nllb}{NLLB family}. These are highly multilingual encoder--decoder Transformer models trained on more than 200 languages. They come in different sizes (from 600M to 54B parameters). We evaluate models up to 3.3B parameters due to computational budget constraints. We also include two variants distilled from their biggest MoE 54B model via Word-level Knowledge Distillation \cite{kim2016sequence}.

    \item \textbf{Llama} \cite{grattafiori2024llama}: 
    We evaluate the instruction-tuned variant \href{https://huggingface.co/meta-llama/Llama-3.1-8B-Instruct}{Llama-3.1-8B-Instruct} via zero-shot prompting. These general-purpose decoder-only LLM is not specifically optimized for translation but has shown competitive performance in zero-shot and few-shot settings \cite{kocmi2025findings}.

    \item \textbf{Tower family}: 
    We evaluate two models from the Tower family \cite{alvestower}. The \href{https://huggingface.co/Unbabel/TowerInstruct-7B-v0.2}{Unbabel/TowerInstruct-7B-v0.2} model is based on Llama 2 \cite{touvron2023llama} and further trained with continued pre-training and supervised fine-tuning for 10 high-resource languages and translation-specific tasks.
    We also evaluate Tower-Plus \cite{rei2025tower}, the model \href{https://huggingface.co/Unbabel/Tower-Plus-9B}{Unbabel/Tower-Plus-9B}, a newer variant trained on Gemma 2 \cite{team2024gemma} optimized for a broader set of multilingual and instruction-following tasks, covering 27 languages.
    
\end{itemize}

All the LLMs are of similar size, between 7B and 9B parameters.

\paragraph{Evaluation Data}

For evaluation, we use the Flores+ benchmark \cite{goyal2022flores}. Flores+ provides professionally translated, multi-domain test sets across a large number of languages, including Esperanto. We evaluate all supported language pairs in both translation directions. We discuss the limitations of using Flores+ in our Limitations Section.

\subsection{MT Development}

In addition to benchmarking existing systems, we train our own open MT models for Esperanto.

\paragraph{Training}
 We train separate bilingual models for each translation direction (into and from Esperanto), following common practices in low-resource MT \cite{haddow2022survey}. We adopt two complementary strategies.

First, we train standard encoder--decoder Transformer models from scratch using Marian \cite{junczys2018marian}, without relying on pretrained models. We experiment with two configurations: Transformer-base (60.6M parameters, \citet{vaswani2017attention}) and Transformer-tiny (17.4M parameters,  \citet{bogoychev-etal-2020-edinburghs}), allowing us to analyze performance–efficiency trade-offs in resource-constrained environments.

Second, we perform supervised fine-tuning of \href{https://huggingface.co/meta-llama/Llama-3.1-8B-Instruct}{Llama-3.1-8B-Instruct}. We apply parameter-efficient fine-tuning using LoRA \cite{hu2022lora}, with rank $r=16$ and scaling factor $\alpha=32$. \textcolor{black}{These hyperparameters were adopted directly from \citet{obrien-etal-2025-dochplt}}. Fine-tuning is conducted using the \texttt{open-instruct} toolkit.\footnote{\url{https://github.com/allenai/open-instruct}}
We adopt an instruction-style prompting format that includes Flores-like language tags (see Figure~\ref{fig:prompt-example}). The model is trained to generate the target translation directly after the instruction prompt.

\begin{figure}[h]
    \centering
    \begin{tcolorbox}[colback=mygreen!35, colframe=myred!65, boxrule=0.5pt]
\small
Translate the text from English (eng\_Latn) into Esperanto (epo\_Latn):

\vspace{5mm}
Majorcan cuisine, like that of similar zones in the Mediterranean, is based on bread, vegetables and meat (specially pork), and uses olive oil throughout.

\vspace{5mm}
Majorka kuirarto, samkiel tiuj de similaj mediteraneaj regionoj, havas kiel bazon panon, legomojn kaj viandon (precipe el porko) kaj ĝi uzadas olivoleon ĉie.

\end{tcolorbox}
    \caption{Prompt example for fine-tuning Llama-3.1-8B-Instruct}
    \label{fig:prompt-example}
\end{figure}

Details of training hyperparameters and Transformer architectures can be found in Appendix \ref{appen:hyperparam}.

We also experimented with fine-tuning NLLB models. However, under our multilingual setup and available data scale, we did not observe improvements over the base models. \textcolor{black}{More details about our failed setup can be found in Appendix~\ref{appen:nllb}.}

\begin{table*}[ht!]
    \centering
    \adjustbox{width=0.85\linewidth}{
    \begin{tabular}{lccccccc}
\toprule
 & eo-en & eo-es & eo-ca & en-eo & es-eo & ca-eo  \\
 \midrule
\rowcolor{gray!15} Rule-based MT & & & & & & \\
\hspace{5mm}  Apertium & \cellcolor{myred!45}47.03 & - & - & 50.82 & 42.68 & 45.44\\
\rowcolor{gray!15}Neural MT & & && & &  \\
\hspace{5mm} NLLB-200-distilled-600M & 65.08 & 48.06 & 52.86 & 58.61 & 48.32 & 52.26 \\
\hspace{5mm} NLLB-200-1.3B & 66.35 & 49.36 & 55.15 & 59.51 & 49.22 & 51.97 \\
\hspace{5mm} NLLB-200-distilled-1.3B & 66.81 & 49.64 &  55.60 &  \cellcolor{mygreen!45}60.04 & 49.34 & 51.52 \\
\hspace{5mm} NLLB-200-3.3B & \cellcolor{mygreen!45} 67.27 &  \cellcolor{mygreen!45}49.68 & \cellcolor{mygreen!45}56.19 & 59.91 &  \cellcolor{mygreen!45}49.57 &  \cellcolor{mygreen!45}52.70 \\
\rowcolor{gray!15}General-purpose LLMs & && & &  & \\
\hspace{5mm} Llama-3.1-8B-Instruct & 62.94 & 45.87 & 48.86 & 55.39 & 44.78 & 49.59 \\
\rowcolor{gray!15}MT-tuned LLMs & & && & &  \\
\hspace{5mm} TowerInstruct-7B-v0.2 & 51.79 &  \cellcolor{myred!45}38.86 &  \cellcolor{myred!45}8.22 &  \cellcolor{myred!45}28.27 &  \cellcolor{myred!45}25.65 &  \cellcolor{myred!45}23.97 \\
\hspace{5mm}  Tower-Plus-9B & 64.63 & 47.66 & 49.02 & 47.02 & 40.38 & 42.99 \\
\bottomrule
\vspace{5mm} \\
\toprule
 & eo-en & eo-es & eo-ca & en-eo & es-eo & ca-eo  \\
 \midrule
\rowcolor{gray!15}Neural MT from Scratch & & & & & & \\
\hspace{5mm} Transformer-base (60.6M) & \cellcolor{mygreen!45}61.33 & \cellcolor{mygreen!45}46.73 & \cellcolor{mygreen!45}53.27 & \cellcolor{mygreen!45}55.11 & 45.97 & \cellcolor{myred!45}48.73 \\
\hspace{5mm} Transformer-tiny (17.4M) & \cellcolor{myred!45}57.69 & \cellcolor{myred!45}45.16 & \cellcolor{myred!45}49.02 & 54.05 & \cellcolor{myred!45}45.33 & 49.57 \\
\rowcolor{gray!15}Fine-tuned General-purpose LLMs & & & & & & \\
\hspace{5mm}  Llama-3.1-8B-Instruct-FT & 61.14 & 45.56 & 49.47 & \cellcolor{myred!45}52.90 & \cellcolor{mygreen!45}46.64 & \cellcolor{mygreen!45}50.35 \\
\bottomrule
    \end{tabular}
    }
    \caption{ChrF++ scores for our benchmarked (above) and trained models (below).  \colorbox{mygreen!45}{Best} and
\colorbox{myred!45}{worst} scores are highlighted for each language direction.}
    \label{tab:results}
\end{table*}

\paragraph{Data}

We use the Tatoeba Challenge data \cite{tiedemann2020tatoeba}, a deduplicated aggregation of parallel corpora from the OPUS repository \cite{tiedemann2024democratizing}. Table \ref{tab:data} shows an overview of the data used.

Data cleaning is performed using OpusFilter \cite{aulamo2020opusfilter}. Our filtering pipeline includes:

\begin{itemize}
    \item Length filtering: minimum 3 tokens, maximum 100 tokens.
    \item Length ratio filtering: maximum ratio of 2 between source and target.
    \item Removal of sentences containing words longer than 40 characters.
    \item Removal of HTML tags.
    \item Language identification filtering with langid.py \cite{lui-baldwin-2012-langid}.
    \item Restriction to Latin alphabet characters.
\end{itemize}

For training the Transformer models, we subsample the English data to 5M to have a more balanced training set.
For LLM fine-tuning, we subsample the data to control training cost. We use 100k sentence pairs per language direction.
We use FLORES+ for both development and evaluation.

\paragraph{Vocabulary} For LLaMA fine-tuning, we use the original LLaMA tokenizer and introduce an additional padding token. For Marian models, we train a multilingual SentencePiece \cite{kudo-richardson-2018-sentencepiece} vocabulary with 32k merge operations, learned jointly over the four languages in our setup. The vocabulary is trained on a balanced 50k-sentence sample per language to avoid dominance of higher-resource languages.

\subsection{Evaluation Metrics}
We report both surface-level and neural evaluation metrics. We compute BLEU \cite{papineni-etal-2002-bleu} and ChrF++ \cite{popovic-2017-chrf},  as n-gram overlap–based measures.  In addition, we report neural metrics, namely COMET\footnote{We use the model \href{https://huggingface.co/Unbabel/wmt22-comet-da}{Unbabel/wmt22-comet-da}.} \cite{rei2022comet} and MetricX\footnote{We use the model \href{https://huggingface.co/google/metricx-24-hybrid-large-v2p6}{google/metricx-24-hybrid-large-v2p6}.} \cite{juraska-etal-2024-metricx}. Neither COMET nor MetricX include Esperanto in their fine-tuning stages. As a result, their scores should be interpreted with caution, as their calibration for Esperanto may be suboptimal. Following recent findings in shared tasks \cite{lavie-etal-2025-findings}, we adopt ChrF++ as our primary metric, as it has been shown to correlate more strongly with human judgments than BLEU, especially for morphologically rich or lower-resource languages.

\section{Results}
\label{sec:results}
Table~\ref{tab:results} reports the ChrF++ scores for both benchmarked models and models trained in this work. Appendix \ref{appen:metrics} shows the scores for BLEU, COMET and MetricX. In general, the four metrics follow similar trends, with similar top- and bottom-performing systems. 
Across systems, translating from Esperanto generally yields higher scores than translating into Esperanto. This asymmetry likely reflects the richer training data available for English, Spanish, and Catalan compared to Esperanto. Consequently, models appear to encode stronger representations for the high-resource languages, while Esperanto generation is more challenging.

%

\begin{figure}[t!]
    \centering
    
    \begin{subfigure}{0.48\linewidth}
        \centering
        \includegraphics[width=\linewidth]{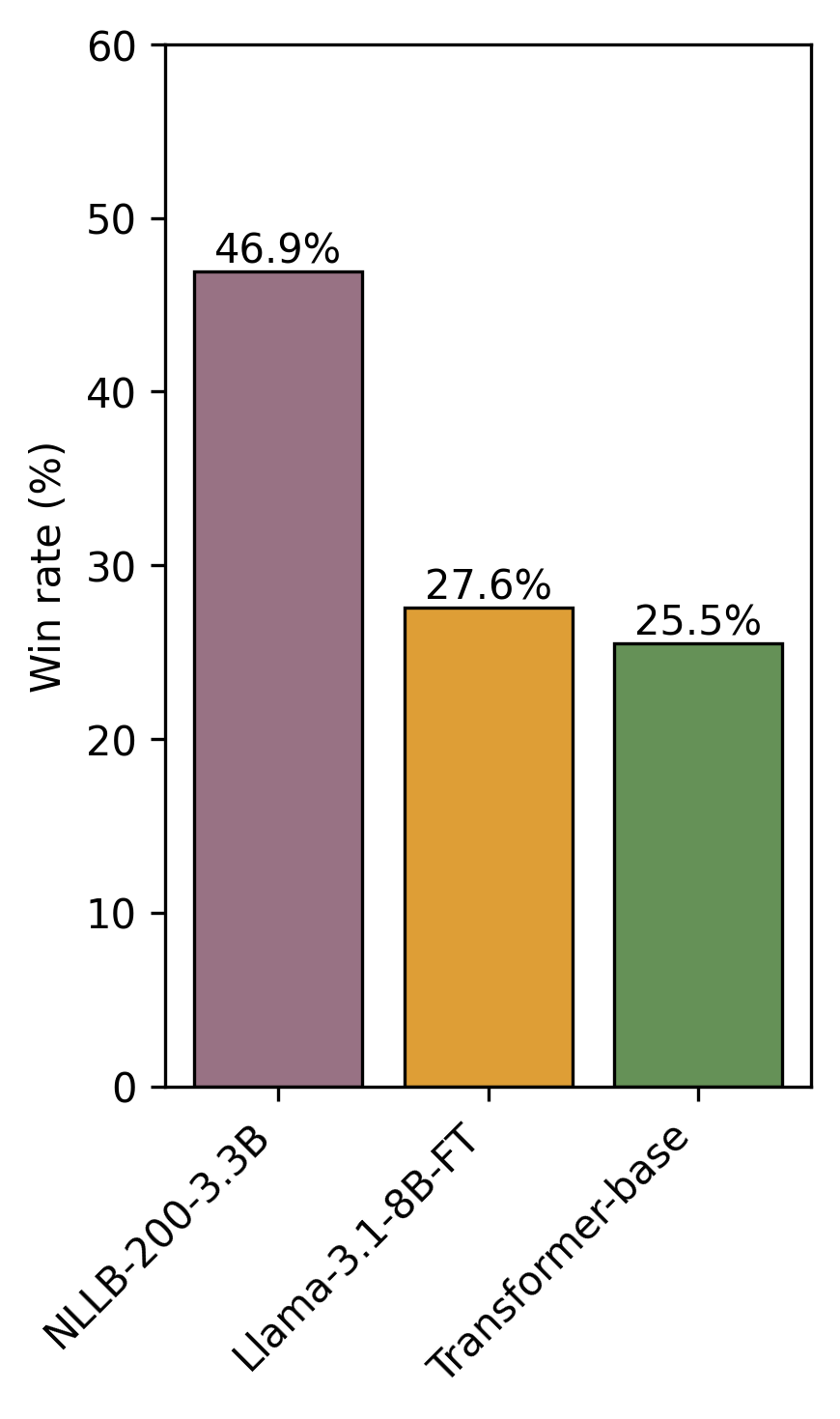}
        \caption{Spanish → Esperanto}
        \label{fig:es-eo}
    \end{subfigure}
    \hfill
    \begin{subfigure}{0.48\linewidth}
        \centering
        \includegraphics[width=\linewidth]{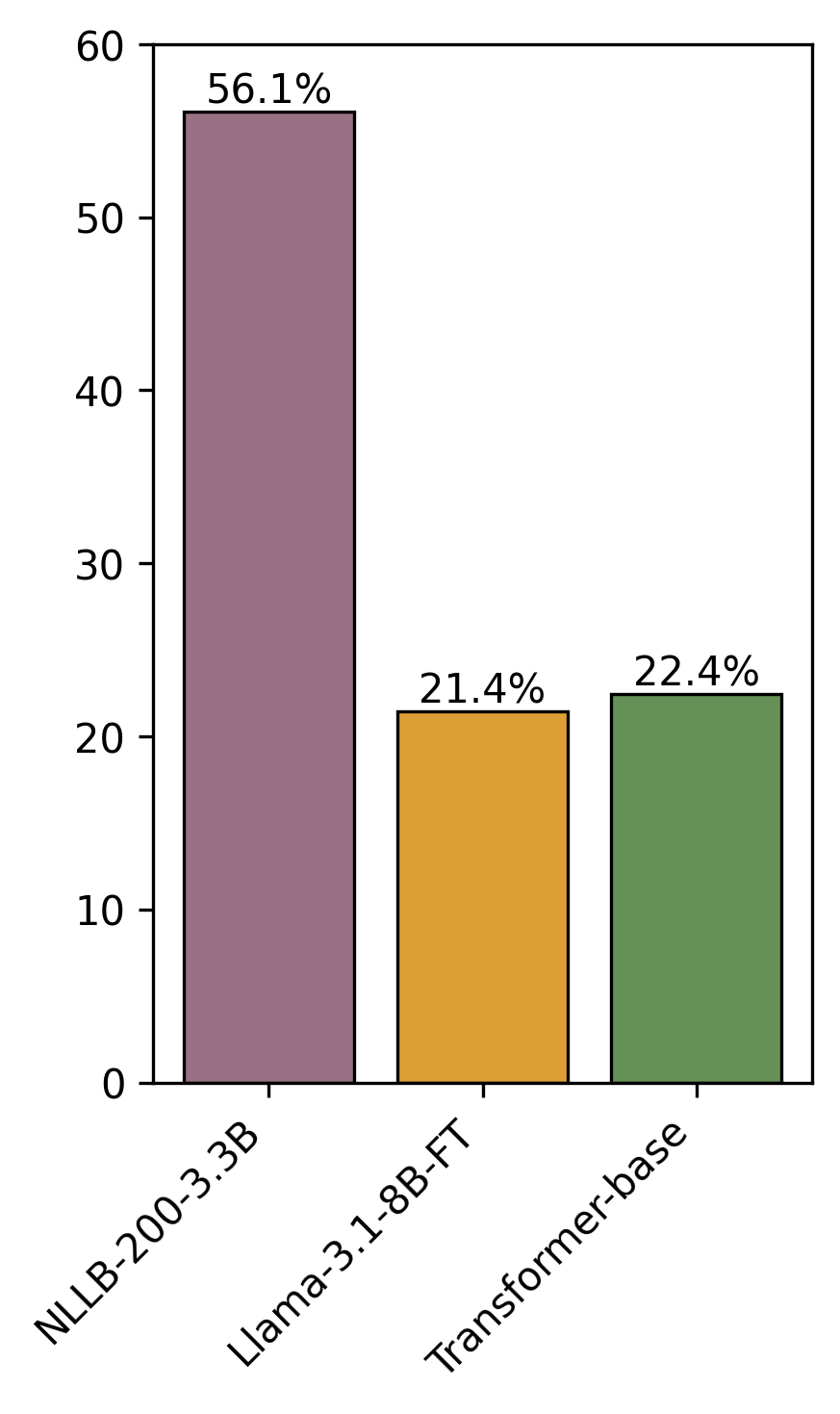}
        \caption{Esperanto → Spanish}
        \label{fig:eo-es}
    \end{subfigure}
    
    \caption{Human evaluation win rates for both translation directions.}
    \label{fig:winrates-both}
\end{figure}

\subsection{Benchmarked Models} The NLLB family consistently outperforms all other models by a clear margin. Performance increases with model size, with NLLB-200-3.3B achieving the highest scores in most directions. The distilled 1.3B variant performs competitively, even slightly surpassing the 3.3B model in one direction (en–eo). 
As hypothesized, LLMs have sufficient pretraining information to understand and produce Esperanto. However, they underperform when compared to dedicated MT systems. While Llama-3.1-8B-Instruct produces reasonable results, both Tower variants lag considerably behind. In particular, TowerInstruct-7B-v0.2 exhibits extremely low scores in several directions, suggesting that its instruction tuning on specific high-resource languages may have negatively affected its general translation capabilities.
\textcolor{black}{A qualitative inspection of its outputs showed that the model often failed to generate Esperanto or Catalan. While it performed somewhat better in directions involving English and Spanish, it still struggled to maintain the requested target language, frequently mixing languages or producing malformed output.} 
Tower-Plus-9B performs more competitively but still falls short of NLLB and strong neural baselines. The model achieves higher performance when translating from Esperanto than when translating into Esperanto. Rule-based Apertium performs substantially worse than neural approaches overall, though, surprisingly, it still surpasses the Tower variants in several directions.

\subsection{Trained Models} Among the models trained in this work, fine-tuning Llama-3.1-8B-Instruct yields only modest improvements. Gains are more noticeable when generating Esperanto, especially for Catalan and Spanish as a source. Performance into English decreases slightly. This is consistent with the expectation that the base model already possesses strong English representations due to extensive pretraining.
Despite its substantially smaller size, the Transformer-base model performs comparably to Llama-3.1-8B-Instruct-FT across most directions, surpassing it in 4 out of 6 pairs. Notably, the Transformer-tiny model achieves surprisingly competitive results, particularly given its limited parameter count. This suggests that compact, task-specific architectures remain strong contenders in low-resource multilingual settings.

\begin{table}[t!]
    \centering
    \adjustbox{width=\linewidth}{
    \begin{tabular}{lrrrr}
    \toprule
    \rowcolor{gray!15}\multicolumn{5}{c}{Spanish $\rightarrow$ Esperanto} \\
    Metric & $\tau$ (mean) & $\tau$ (pooled) & P-value & Accuracy  \\
    \midrule
BLEU & \colorbox{myred!45}{0.061} & \colorbox{myred!45}{-0.023}& 0.610 & \colorbox{myred!45}{0.522} \\
ChrF++ & 0.116 & -0.049 & 0.276  & 0.558 \\
COMET & 0.191 & -0.110 & \textbf{0.015*} & 0.595 \\
MetricX & \colorbox{mygreen!45}{0.327} & \colorbox{mygreen!45}{0.165} & \textbf{2.6x10-4*} & \colorbox{mygreen!45}{0.663} \\
\bottomrule
\vspace{5mm} \\
\toprule
\rowcolor{gray!15}\multicolumn{5}{c}{Esperanto $\rightarrow$ Spanish} \\
Metric & $\tau$ (mean) & $\tau$ (pooled) & P-value & Accuracy  \\
\midrule
BLEU & \colorbox{myred!45}{0.361} & \colorbox{myred!45}{-0.123} & \textbf{6.24x10-3*} & \colorbox{myred!45}{0.668} \\
ChrF++ & 0.347 & -0.146 &  \textbf{1.18x10-3*} & 0.672 \\
COMET & \colorbox{mygreen!45}{0.422} & -0.177 &  \textbf{8.6x10-5*} & \colorbox{mygreen!45}{0.711} \\
MetricX & 0.415 & \colorbox{mygreen!45}{0.200} &  \textbf{9.5x10-6*} & 0.708 \\
    \bottomrule
    \end{tabular}}
    \caption{Correlation between human judgments and automatic evaluation metrics. Results are reported as mean Kendall’s $\tau$ over per-sentence rankings, pooled Kendall’s $\tau$ over metric scores with the corresponding p-value, and pairwise accuracy. \colorbox{mygreen!45}{Highest} and \colorbox{myred!45}{lowest} values are highlighted. Statistically significant values ($p < 0.05$) are bolded and marked with an asterisk (*).}
    \label{tab:corr}
\end{table}

\section{Human Evaluation}
\label{sec:human}
To complement the automatic evaluation, we conducted a human assessment of translation quality for the Spanish--Esperanto language pair. We randomly sampled 100 source sentences and extracted the corresponding translations produced by the top three models for each architecture, namely, Transformer-base, NLLB-200-3.3B, and Llama-3.1-8B-Instruct-FT. For every source sentence, the three system outputs were presented in randomized order to a human annotator, who was asked to rank them by selecting the best and the worst translation\footnote{Two human annotators were employed, one for each translation direction. L1: Catalan, Spanish. L2: Esperanto.}. An optional comment field allowed the annotator to justify their choice or note specific errors briefly, without any specific guidelines. This pairwise ranking setup \cite{laubli-etal-2018-machine} enables direct comparison between models while keeping the annotation procedure simple and intuitive.
The annotation guidelines can be found in Appendix \ref{appen:guidelines}.

\subsection{Results} Figure \ref{fig:winrates-both} shows the win rates achieved by the three models on the human evaluation task for both language directions\footnote{We removed two sentences for each translation direction where two of translations resulted in a tie.}. 
These results confirm the trends observed in the automatic metrics.
In both cases, NLLB stands out as the clear winner, selected as the best translation around 50\% of the time. The other two systems perform considerably worse and at a similar level. 
The advantage of NLLB is particularly pronounced in the Esperanto $\rightarrow$ Spanish direction, which suggests once more that generating Esperanto is generally more challenging across models.
However, these results reflect relative differences between systems rather than absolute translation quality, and even the best-performing system still produces noticeable errors.

\subsection{Metric Correlations with Human Judgements}
We compute correlations of the human judgments with the four automatic metrics with Kendall's $\tau$ \cite{machacek-bojar-2013-results,deutsch-etal-2023-ties}. We use three complementary measures: (i) Kendall’s $\tau$ over model rankings, computed per sample and averaged, to measure how well each metric reproduced the human ordering of translations; (ii) pooled Kendall’s $\tau$ over model scores, computed across all translations, to measure the global monotonic relationship between metric scores and human quality; and (iii) pairwise accuracy, measuring how often a metric correctly identified the better translation in each pairwise comparison.

Table~\ref{tab:corr} shows the results for both translation directions. The results reveal a consistent hierarchy of metric quality across directions. MetricX and COMET achieve the strongest agreement with human judgments, with MetricX performing best overall and COMET showing comparable performance, particularly in the Esperanto$\rightarrow$Spanish direction. These metrics show statistically significant agreement in both directions. In contrast, ChrF++ shows weaker agreement with human rankings, while BLEU performs close to random in the Spanish$\rightarrow$Esperanto direction but shows moderate correlation in the other. Overall, the learned metrics correlate substantially better with human judgments than traditional n-gram-based metrics, even though they have not been directly exposed to Esperanto in the fine-tuning stage.

\subsection{Qualitative Error Analysis} We summarize the free comments from the human evaluation and provide qualitative insights into recurring error patterns.
Appendix \ref{appen:error} provides illustrative examples.


For the Transformer-base model, we observe a range of lexical and grammatical errors. The model occasionally produces non-existent words or leaves source words untranslated. Grammatical problems include incorrect verb forms, agreement errors between articles and nouns, and missing verbs. The model also struggles with named entity translation and occasionally mistranslates relatively simple lexical items.
In contrast, the NLLB-200-3.3B model generally produces fluent and semantically adequate translations. Most errors appear to stem from the compositional nature of Esperanto word formation. For example, \textit{nekredantoj} is rendered as \textit{incrédulos}, while \textit{neĝtabulo} and \textit{flugaparatoj} are translated compositionally as \textit{tabla de nieve} and \textit{aparatos de vuelo}. Although these translations remain understandable, they are less appropriate than conventional equivalents (\textit{snowboard}, \textit{aviones}). In one case, it omits semantically relevant information.
Finally, the Llama-3.1-8B-Instruct-FT model exhibits the widest range of error types. The model recurrently adds, omits, or invents information. In addition, it sometimes introduces English words into the output and produces grammatical errors of varying severity, including agreement errors and semantic distortions such as incorrectly assigning the subject.

These error patterns are in line with our expectations. The Transformer-base model tends to produce accurate but literal translations, it sometimes generates unusual constructions and wrong named entity translations. Both NLLB-200-3.3B and  Llama-3.1-8B-Instruct-FT produce fluent output; however, NLLB-200-3.3B can occasionally be overly literal, while Llama more frequently modifies or invents information, which may pose risks in real-world deployment.

\section{Discussion}
In this section, we discuss the main findings of our experiments and their implications for Esperanto translation and, more broadly, low-resource MT.

\subsection{NLLB Remains a Strong Baseline}
Although the NLLB model family is now several years old, it remains by far the strongest system in our experiments. We hypothesize that this is due to NLLB’s highly multilingual training, which allows it to benefit from transfer learning, as well as its explicit training for translation.
This finding is consistent with recent work on low-resource MT, where NLLB continues to outperform a wide range of alternative approaches \cite{de-gibert-etal-2025-scaling,scalvini-etal-2025-rethinking,tapo-etal-2025-bayelemabaga,Aycock24}. The distilled 1.3B model performs slightly better than its non-distilled counterpart. Even the smallest model in the family achieves strong results across language directions. These observations support a clear practical recommendation: for low-resource MT, NLLB should be considered the default choice, with model size selected according to available computational resources, prioritizing distilled variants.

\subsection{General-Purpose LLMs Outperform MT-Tuned LLMs}
The MT-tuned LLMs evaluated in this work, which are specifically designed for translation tasks, consistently underperform the general-purpose LLM baseline. In particular, TowerInstruct-7B-v0.2, which was fine-tuned primarily on 10 high-resource languages, is largely unable to produce meaningful output beyond English; even though its fine-tuning also includes Spanish. Tower-Plus-9B performs similarly to Llama-3.1-8B-Instruct-FT but is poor at Esperanto generation. 
These findings suggest that, for low-resource scenarios where NLLB coverage is unavailable, general-purpose LLMs appear to be a more reliable choice than translation-specialized LLMs, since language-specific fine-tuning may reduce their general translation abilities.


\subsection{Data-Hungry vs.\ Compute-Hungry Models}
In our training setup, we compare a fine-tuned Llama model with encoder–decoder models. When only limited training data is available, fine-tuning a pretrained Llama model is the most effective approach: even with as little as 100k sentence pairs per language direction, fine-tuning yields competitive results. However, under constrained computational budgets, training models from scratch becomes an attractive alternative, provided that sufficient training data is available.
Our smallest Transformer models are more than 500 times smaller than Llama-3.1-8B-Instruct-FT, yet achieve comparable and, in one case (en-eo), superior performance. This result highlights a broader tendency in the field toward increasingly large architectures, even in scenarios where smaller models can achieve similar quality. Moreover, our compact models can run efficiently on standard CPUs, making them suitable for deployment on personal devices. This aligns well with the principles of Esperanto as a language intended to facilitate universal communication, as lightweight models lower the computational barriers to MT. To support accessibility and reproducibility, we release our Transformer models on HuggingFace.

\subsection{Neural Metrics are Effective in Zero-Shot Settings}
Our analysis in Table~\ref{tab:corr} shows that learned metrics correlate substantially better with human judgments than traditional metrics. Neither metric has been explicitly fine-tuned on Esperanto data. MetricX is based on mT5 \cite{xue-etal-2021-mt5}, while COMET builds on XLM-RoBERTa-base \cite{conneau2020unsupervised}; both pre-trained models include Esperanto and Spanish in their multilingual training data. However, the subsequent fine-tuning of these metrics on WMT datasets \cite{kocmi2025findings} does not involve Esperanto. The strong performance observed in our experiments, therefore, suggests that transfer learning enables neural metrics to generalize effectively to previously unseen language pairs.
We hypothesize that the strong performance of these metrics may be partly explained by the linguistic characteristics of Esperanto, which is largely derived from Romance languages. A systematic evaluation across a broader set of low-resource languages would be necessary to assess the generalization of these findings.

\section{Conclusions}
\label{sec:conclusion}
Esperanto is a widely used conlang whose community aligns with ideals of linguistic and technological sovereignty. 
We systematically study Esperanto translation with a particular emphasis on open models. We evaluate a range of existing systems and develop compact models of our own, demonstrating that high translation quality can be achieved with remarkably small architectures. Our results confirm that NLLB remains the strongest overall system, while general-purpose LLMs perform similarly to our task-specific Transformer models despite being orders of magnitude larger. The efficiency of these smaller models makes them faster, more accessible, and environmentally sustainable, aligning closely with the practical and ideological goals of the Esperanto community. Ensuring the continued development of free MT systems is essential for maintaining a digital linguistic infrastructure that can be governed, audited, and adapted by members of the community who rely on it. This work represents a first step toward that goal.

For future work, we are interested in studying whether Esperanto is inherently easier to model than natural languages due to its regular morphological and syntactic structure, following work by \citet{ploeger-etal-2025-cross}. Another important direction would be to revisit the original idea of Esperanto as an interlingua for pivot-based MT.
Finally, given the availability of rule-based resources in Apertium, future work could explore hybrid approaches that leverage RBMT knowledge \cite{de-gibert-etal-2024-hybrid}.

\section*{Limitations}
\label{sec:limits}
We are aware of our limited evaluation setup covering only one test set, Flores+. Flores+ was created by translating directly from English, which may introduce biases that affect evaluation. However, Esperanto is not present in any other MT benchmark. To compensate for this, we perform human evaluation and report a diverse set of evaluation metrics.
Furthermore, our human evaluation is limited to 100 samples per language direction and one annotator per direction due to a lack of resources.

\section*{Ethical considerations}
All annotators are authors of this paper, and the total time spent on individual annotations did not exceed four hours.

\section*{Acknowledgments}
\textcolor{black}{We thank Seth Aycock and Joseph Attieh for their insightful feedback and valuable comments.}

\textcolor{black}{This project has received funding from the Digital Europe programme of the European Union under Grant No.~101195233.
The contents of this publication are the sole responsibility of its authors and do not necessarily reflect the opinion of the European Union.}

\section{Bibliographical References}

\bibliographystyle{lrec2026-natbib}
\bibliography{lrec2026-example}

@inproceedings{minnaja-paccagnella-2000-part,
    title = "A part-of-speech tagger for {E}speranto oriented to {MT}",
    author = "Minnaja, Carlo  and
      Paccagnella, Laura",
    booktitle = "Proceedings of the International Conference on Machine Translation and Multilingual Applications in the new Millennium: MT 2000",
    month = nov # " 20-22",
    year = "2000",
    address = "University of Exeter, UK",
    url = "https://aclanthology.org/2000.bcs-1.13/"
}

@inproceedings{witkam-1984-distributed,
    title = "Distributed language translation, another {MT} system",
    author = "Witkam, A. P. M.",
    booktitle = "Proceedings of the International Conference on Methodology and Techniques of Machine Translation: Processing from words to language",
    month = feb # " 13-15",
    year = "1984",
    address = "Cranfield University, UK",
    url = "https://aclanthology.org/1984.bcs-1.34/"
}

@article{neijt1986esperanto,
  title={Esperanto as the focal point of machine translation},
  author={Neijt, A},
  journal={Multilingua},
  volume={5},
  number={1},
  pages={9--13},
  year={1986},
  publisher={Walter de Gruyter \& Co. Hawthorne, NJ, USA}
}

@inproceedings{bick2011wikitrans,
  title={WikiTrans: the English Wikipedia in Esperanto},
  author={Bick, Eckhard},
  booktitle={Constraint Grammar Applications, Workshop Proceedings at Nodalida},
  volume={14},
  pages={8--16},
  year={2011}
}

@inproceedings{karlsson1990constraint,
  title={Constraint grammar as a framework for parsing running text},
  author={Karlsson, Fred},
  booktitle={COLING 1990 Volume 3: Papers presented to the 13th International Conference on Computational Linguistics},
  year={1990}
}

@inproceedings{poncelas2020using,
  title={Using Multiple Subwords to Improve English-Esperanto Automated Literary Translation Quality},
  author={Poncelas, Alberto and Buts, Jan and Hadley, James and Way, Andy},
  booktitle={Proceedings of the 3rd Workshop on Technologies for MT of Low Resource Languages},
  pages={108--117},
  year={2020}
}

@incollection{schubert2002esperanto,
  title={Esperanto as an intermediate language for machine translation},
  author={Schubert, Klaus},
  booktitle={Computers in translation},
  pages={98--115},
  year={2002},
  publisher={Routledge}
}

@inproceedings{manaris2006investigating,
  title={Investigating Esperanto's Statistical Proportions Relative to other Languages using Neural Networks and Zipf's Law.},
  author={Manaris, Bill Z and Pellicoro, Luca and Pothering, George J and Hodges, Harland},
  booktitle={Artificial Intelligence and Applications},
  pages={102--108},
  year={2006}
}

@article{boddington2004evaluation,
  title={Evaluation of an Esperanto-Based Interlingua Multilingual Survey Form Machine Translation Mechanism Incorporating a Sublanguage Translation Methodolgy},
  author={Boddington, Richard},
  year={2004},
  publisher={Edith Cowan University}
}

@article{orlova2015esperus,
  title={Esperus: the First Step to Build a Statistical Machine. Translation System for Esperanto and Russian Languages},
  author={Orlova, Darja},
  journal={AINL FRUCT, Saint Petersburg, Russia},
  year={2015}
}

@inproceedings{oya-2025-ud,
    title = "{UD} Treebanks for {E}speranto as a natural language",
    author = "Oya, Masanori",
    editor = {Bouma, Gosse  and
      {\c{C}}{\"o}ltekin, {\c{C}}a{\u{g}}r{\i}},
    booktitle = "Proceedings of the Eighth Workshop on Universal Dependencies (UDW, SyntaxFest 2025)",
    month = aug,
    year = "2025",
    address = "Ljubljana, Slovenia",
    publisher = "Association for Computational Linguistics",
    url = "https://aclanthology.org/2025.udw-1.3/",
    pages = "22--29",
    ISBN = "979-8-89176-292-3"
}

@article{FIEDLER2018166,
title = {Linguistic and pragmatic influence of English: Does Esperanto resist it?},
journal = {Journal of Pragmatics},
volume = {133},
pages = {166-178},
year = {2018},
issn = {0378-2166},
doi = {https://doi.org/10.1016/j.pragma.2018.05.007},
url = {https://www.sciencedirect.com/science/article/pii/S0378216617305180},
author = {Sabine Fiedler}
}

@inproceedings{franco-sabaris-etal-2001-multilingual,
    title = "Multilingual authoring through an artificial language",
    author = "Franco Sabar{\'i}s, Marcos  and
      Rojas Alonso, Jos{\'e} Luis  and
      Dafonte, C.  and
      Arcay, B.",
    editor = "Maegaard, Bente",
    booktitle = "Proceedings of Machine Translation Summit VIII",
    month = sep # " 18-22",
    year = "2001",
    address = "Santiago de Compostela, Spain",
    url = "https://aclanthology.org/2001.mtsummit-papers.19/"
}

@article{gobbo2015machine,
  title={Machine translation as a complex system: The role of Esperanto},
  author={Gobbo, Federico},
  journal={Interdisciplinary Description of Complex Systems: INDECS},
  volume={13},
  number={2},
  pages={264--274},
  year={2015},
  publisher={Hrvatsko interdisciplinarno dru{\v{s}}tvo}
}

@article{couturat1903histoire,
  title={Histoire de la langue universelle},
  author={Couturat, L},
  journal={Hildesheim, Z{\"u}rich, \& New York: Olms},
  year={1903}
}

@inproceedings{tiedemann2020tatoeba,
  title={The Tatoeba Translation Challenge--Realistic Data Sets for Low Resource and Multilingual MT},
  author={Tiedemann, J{\"o}rg},
  booktitle={Proceedings of the Fifth Conference on Machine Translation},
  pages={1174--1182},
  year={2020}
}

@article{tiedemann2024democratizing,
  title={Democratizing neural machine translation with OPUS-MT},
  author={Tiedemann, J{\"o}rg and Aulamo, Mikko and Bakshandaeva, Daria and Boggia, Michele and Gr{\"o}nroos, Stig-Arne and Nieminen, Tommi and Raganato, Alessandro and Scherrer, Yves and V{\'a}zquez, Ra{\'u}l and Virpioja, Sami},
  journal={Language Resources and Evaluation},
  volume={58},
  number={2},
  pages={713--755},
  year={2024},
  publisher={Springer}
}

@inproceedings{aulamo2020opusfilter,
  title={OpusFilter: A configurable parallel corpus filtering toolbox},
  author={Aulamo, Mikko and Virpioja, Sami and Tiedemann, J{\"o}rg},
  booktitle={2020 Annual Conference of the Association for Computational Linguistics},
  pages={150--156},
  year={2020},
  organization={The Association for Computational Linguistics}
}

@article{goyal2022flores,
  title={The flores-101 evaluation benchmark for low-resource and multilingual machine translation},
  author={Goyal, Naman and Gao, Cynthia and Chaudhary, Vishrav and Chen, Peng-Jen and Wenzek, Guillaume and Ju, Da and Krishnan, Sanjana and Ranzato, Marc’Aurelio and Guzm{\'a}n, Francisco and Fan, Angela},
  journal={Transactions of the Association for Computational Linguistics},
  volume={10},
  pages={522--538},
  year={2022},
  publisher={MIT Press One Broadway, 12th Floor, Cambridge, Massachusetts 02142, USA~…}
}

@inproceedings{koehn2007moses,
  title={Moses: Open source toolkit for statistical machine translation},
  author={Koehn, Philipp and Hoang, Hieu and Birch, Alexandra and Callison-Burch, Chris and Federico, Marcello and Bertoldi, Nicola and Cowan, Brooke and Shen, Wade and Moran, Christine and Zens, Richard and others},
  booktitle={Proceedings of the 45th annual meeting of the association for computational linguistics companion volume proceedings of the demo and poster sessions},
  pages={177--180},
  year={2007}
}

@article{hu2022lora,
  title={Lora: Low-rank adaptation of large language models.},
  author={Hu, Edward J and Shen, Yelong and Wallis, Phillip and Allen-Zhu, Zeyuan and Li, Yuanzhi and Wang, Shean and Wang, Lu and Chen, Weizhu and others},
  journal={ICLR},
  volume={1},
  number={2},
  pages={3},
  year={2022}
}

@inproceedings{junczys2018marian,
  title={Marian: Fast Neural Machine Translation in C++},
  author={Junczys-Dowmunt, Marcin and Grundkiewicz, Roman and Dwojak, Tomasz and Hoang, Hieu and Heafield, Kenneth and Neckermann, Tom and Seide, Frank and Germann, Ulrich and Aji, Alham Fikri and Bogoychev, Nikolay and others},
  booktitle={Proceedings of ACL 2018, System Demonstrations},
  pages={116--121},
  year={2018}
}

@inproceedings{de-gibert-etal-2024-hybrid,
    title = "Hybrid Distillation from {RBMT} and {NMT}: {H}elsinki-{NLP}{'}s Submission to the Shared Task on Translation into Low-Resource Languages of {S}pain",
    author = {De Gibert, Ona  and
      Aulamo, Mikko  and
      Scherrer, Yves  and
      Tiedemann, J{\"o}rg},
    editor = "Haddow, Barry  and
      Kocmi, Tom  and
      Koehn, Philipp  and
      Monz, Christof",
    booktitle = "Proceedings of the Ninth Conference on Machine Translation",
    month = nov,
    year = "2024",
    address = "Miami, Florida, USA",
    publisher = "Association for Computational Linguistics",
    url = "https://aclanthology.org/2024.wmt-1.88/",
    doi = "10.18653/v1/2024.wmt-1.88",
    pages = "908--917"
}

@article{blanke2009causes,
  title={Causes of the relative success of Esperanto},
  author={Blanke, Detlev},
  journal={Language Problems and Language Planning},
  volume={33},
  number={3},
  pages={251--266},
  year={2009},
  publisher={John Benjamins}
}

@inproceedings{ploeger-etal-2025-cross,
    title = "A Cross-Lingual Perspective on Neural Machine Translation Difficulty",
    author = {Ploeger, Esther  and
      Bjerva, Johannes  and
      Tiedemann, J{\"o}rg  and
      {\"O}stling, Robert},
    editor = "Haddow, Barry  and
      Kocmi, Tom  and
      Koehn, Philipp  and
      Monz, Christof",
    booktitle = "Proceedings of the Tenth Conference on Machine Translation",
    month = nov,
    year = "2025",
    address = "Suzhou, China",
    publisher = "Association for Computational Linguistics",
    url = "https://aclanthology.org/2025.wmt-1.21/",
    doi = "10.18653/v1/2025.wmt-1.21",
    pages = "340--354",
    ISBN = "979-8-89176-341-8"
}

@article{apertium,
author = {Forcada, Mikel L. and Ginest\'{\i}-Rosell, Mireia and Nordfalk, Jacob and O'Regan, Jim and Ortiz-Rojas, Sergio and P\'{e}rez-Ortiz, Juan Antonio and S\'{a}nchez-Mart\'{\i}nez, Felipe and Ram\'{\i}rez-S\'{a}nchez, Gema and Tyers, Francis M.},
title = {Apertium: a free/open-source platform for rule-based machine translation},
year = {2011},
issue_date = {June      2011},
publisher = {Kluwer Academic Publishers},
address = {USA},
volume = {25},
number = {2},
issn = {0922-6567},
url = {https://doi.org/10.1007/s10590-011-9090-0},
doi = {10.1007/s10590-011-9090-0},
journal = {Machine Translation},
month = jun,
pages = {127–144},
numpages = {18}
}

@article{burghelea2019not,
  title={On Not Being Lost in Translation: Creative Strategies to Approach Multiculturalism in Esperanto},
  author={Burghelea, Manuela},
  journal={J{\k{e}}zyk. Komunikacja. Informacja},
  number={13},
  pages={159--174},
  year={2019},
  publisher={Uniwersytet im. Adama Mickiewicza w Poznaniu}
}

@inproceedings{bick-2025-annotated,
    title = "An Annotated Error Corpus for {E}speranto",
    author = "Bick, Eckhard",
    editor = "Trosterud, Trond  and
      Wiechetek, Linda  and
      Pirinen, Flammie",
    booktitle = "Proceedings of the 9th Workshop on Constraint Grammar and Finite State NLP",
    month = mar,
    year = "2025",
    address = "Tallinn, Estonia",
    publisher = "University of Tartu Library",
    url = "https://aclanthology.org/2025.cgmta-1.1/",
    pages = "1--8",
    ISBN = "978-9908-53-113-7"
}

@inproceedings{bick-2016-morphological,
    title = "A Morphological Lexicon of {E}speranto with Morpheme Frequencies",
    author = "Bick, Eckhard",
    editor = "Calzolari, Nicoletta  and
      Choukri, Khalid  and
      Declerck, Thierry  and
      Goggi, Sara  and
      Grobelnik, Marko  and
      Maegaard, Bente  and
      Mariani, Joseph  and
      Mazo, Helene  and
      Moreno, Asuncion  and
      Odijk, Jan  and
      Piperidis, Stelios",
    booktitle = "Proceedings of the Tenth International Conference on Language Resources and Evaluation ({LREC}'16)",
    month = may,
    year = "2016",
    address = "Portoro{\v{z}}, Slovenia",
    publisher = "European Language Resources Association (ELRA)",
    url = "https://aclanthology.org/L16-1171/",
    pages = "1075--1078"
}

@inproceedings{bick-2020-syntax,
    title = "Syntax and Semantics in a Treebank for {E}speranto",
    author = "Bick, Eckhard",
    editor = "Calzolari, Nicoletta  and
      B{\'e}chet, Fr{\'e}d{\'e}ric  and
      Blache, Philippe  and
      Choukri, Khalid  and
      Cieri, Christopher  and
      Declerck, Thierry  and
      Goggi, Sara  and
      Isahara, Hitoshi  and
      Maegaard, Bente  and
      Mariani, Joseph  and
      Mazo, H{\'e}l{\`e}ne  and
      Moreno, Asuncion  and
      Odijk, Jan  and
      Piperidis, Stelios",
    booktitle = "Proceedings of the Twelfth Language Resources and Evaluation Conference",
    month = may,
    year = "2020",
    address = "Marseille, France",
    publisher = "European Language Resources Association",
    url = "https://aclanthology.org/2020.lrec-1.630/",
    pages = "5120--5127",
    language = "eng",
    ISBN = "979-10-95546-34-4"
}

@article{costa2022no,
  title={No language left behind: Scaling human-centered machine translation},
  author={Costa-Juss{\`a}, Marta R and Cross, James and {\c{C}}elebi, Onur and Elbayad, Maha and Heafield, Kenneth and Heffernan, Kevin and Kalbassi, Elahe and Lam, Janice and Licht, Daniel and Maillard, Jean and others},
  journal={arXiv preprint arXiv:2207.04672},
  year={2022}
}

@article{nllb2024scaling,
  title={Scaling neural machine translation to 200 languages},
  author={{NLLB Team}},
  journal={Nature},
  volume={630},
  number={8018},
  pages={841--846},
  year={2024},
  publisher={Nature Publishing Group UK London}
}

@inproceedings{kim2016sequence,
  title={Sequence-level knowledge distillation},
  author={Kim, Yoon and Rush, Alexander M},
  booktitle={Proceedings of the 2016 conference on empirical methods in natural language processing},
  pages={1317--1327},
  year={2016}
}

@article{grattafiori2024llama,
  title={The llama 3 herd of models},
  author={Grattafiori, Aaron and Dubey, Abhimanyu and Jauhri, Abhinav and Pandey, Abhinav and Kadian, Abhishek and Al-Dahle, Ahmad and Letman, Aiesha and Mathur, Akhil and Schelten, Alan and Vaughan, Alex and others},
  journal={arXiv preprint arXiv:2407.21783},
  year={2024}
}

@inproceedings{alvestower,
  title={Tower: An Open Multilingual Large Language Model for Translation-Related Tasks},
  year={2024},
  author={Alves, Duarte Miguel and Pombal, Jos{\'e} and Guerreiro, Nuno M and Martins, Pedro Henrique and Alves, Jo{\~a}o and Farajian, Amin and Peters, Ben and Rei, Ricardo and Fernandes, Patrick and Agrawal, Sweta and others},
  booktitle={First Conference on Language Modeling}
}

@article{rei2025tower,
  title={Tower+: Bridging generality and translation specialization in multilingual llms},
  author={Rei, Ricardo and Guerreiro, Nuno M and Pombal, Jos{\'e} and Alves, Jo{\~a}o and Teixeirinha, Pedro and Farajian, Amin and Martins, Andr{\'e} FT},
  journal={arXiv preprint arXiv:2506.17080},
  year={2025}
}

@article{occhini2026artificial,
  title={Artificial intelligence is creating a new global linguistic hierarchy},
  author={Occhini, Giulia and Tanaka-Ishii, Kumiko and Barford, Anna and Tikochinski, Refael and Hu, Songbo and Reichart, Roi and Zhou, Yijie and Claus, Hannah and Petti, Ulla and Vuli{\'c}, Ivan and others},
  journal={arXiv preprint arXiv:2602.12018},
  year={2026}
}

@article{haddow2022survey,
  title={Survey of low-resource machine translation},
  author={Haddow, Barry and Bawden, Rachel and Miceli-Barone, Antonio Valerio and Helcl, Jind{\v{r}}ich and Birch, Alexandra},
  journal={Computational Linguistics},
  volume={48},
  number={3},
  pages={673--732},
  year={2022}
}

@inproceedings{obrien-etal-2025-dochplt,
    title = "{D}oc{HPLT}: A Massively Multilingual Document-Level Translation Dataset",
    author = {O{'}Brien, Dayy{\'a}n  and
      Malik, Bhavitvya  and
      de Gibert, Ona  and
      Chen, Pinzhen  and
      Haddow, Barry  and
      Tiedemann, J{\"o}rg},
    editor = "Haddow, Barry  and
      Kocmi, Tom  and
      Koehn, Philipp  and
      Monz, Christof",
    booktitle = "Proceedings of the Tenth Conference on Machine Translation",
    month = nov,
    year = "2025",
    address = "Suzhou, China",
    publisher = "Association for Computational Linguistics",
    url = "https://aclanthology.org/2025.wmt-1.17/",
    doi = "10.18653/v1/2025.wmt-1.17",
    pages = "286--300",
    ISBN = "979-8-89176-341-8"
}

@inproceedings{rei2022comet,
  title={COMET-22: Unbabel-IST 2022 submission for the metrics shared task},
  author={Rei, Ricardo and De Souza, Jos{\'e} GC and Alves, Duarte and Zerva, Chrysoula and Farinha, Ana C and Glushkova, Taisiya and Lavie, Alon and Coheur, Luisa and Martins, Andr{\'e} FT},
  booktitle={Proceedings of the Seventh Conference on Machine Translation (WMT)},
  pages={578--585},
  year={2022}
}

@article{touvron2023llama,
  title={Llama 2: Open foundation and fine-tuned chat models},
  author={Touvron, Hugo and Martin, Louis and Stone, Kevin and Albert, Peter and Almahairi, Amjad and Babaei, Yasmine and Bashlykov, Nikolay and Batra, Soumya and Bhargava, Prajjwal and Bhosale, Shruti and others},
  journal={arXiv preprint arXiv:2307.09288},
  year={2023}
}

@inproceedings{machacek-bojar-2013-results,
    title = "Results of the {WMT}13 Metrics Shared Task",
    author = "Mach{\'a}{\v{c}}ek, Matou{\v{s}}  and
      Bojar, Ond{\v{r}}ej",
    editor = "Bojar, Ondrej  and
      Buck, Christian  and
      Callison-Burch, Chris  and
      Haddow, Barry  and
      Koehn, Philipp  and
      Monz, Christof  and
      Post, Matt  and
      Saint-Amand, Herve  and
      Soricut, Radu  and
      Specia, Lucia",
    booktitle = "Proceedings of the Eighth Workshop on Statistical Machine Translation",
    month = aug,
    year = "2013",
    address = "Sofia, Bulgaria",
    publisher = "Association for Computational Linguistics",
    url = "https://aclanthology.org/W13-2202/",
    pages = "45--51"
}

@inproceedings{bogoychev-etal-2020-edinburghs,
    title = "{E}dinburgh{'}s Submissions to the 2020 Machine Translation Efficiency Task",
    author = "Bogoychev, Nikolay  and
      Grundkiewicz, Roman  and
      Aji, Alham Fikri  and
      Behnke, Maximiliana  and
      Heafield, Kenneth  and
      Kashyap, Sidharth  and
      Farsarakis, Emmanouil-Ioannis  and
      Chudyk, Mateusz",
    editor = "Birch, Alexandra  and
      Finch, Andrew  and
      Hayashi, Hiroaki  and
      Heafield, Kenneth  and
      Junczys-Dowmunt, Marcin  and
      Konstas, Ioannis  and
      Li, Xian  and
      Neubig, Graham  and
      Oda, Yusuke",
    booktitle = "Proceedings of the Fourth Workshop on Neural Generation and Translation",
    month = jul,
    year = "2020",
    address = "Online",
    publisher = "Association for Computational Linguistics",
    url = "https://aclanthology.org/2020.ngt-1.26/",
    doi = "10.18653/v1/2020.ngt-1.26",
    pages = "218--224"
}

@inproceedings{kudo-richardson-2018-sentencepiece,
    title = "{S}entence{P}iece: A simple and language independent subword tokenizer and detokenizer for Neural Text Processing",
    author = "Kudo, Taku  and
      Richardson, John",
    editor = "Blanco, Eduardo  and
      Lu, Wei",
    booktitle = "Proceedings of the 2018 Conference on Empirical Methods in Natural Language Processing: System Demonstrations",
    month = nov,
    year = "2018",
    address = "Brussels, Belgium",
    publisher = "Association for Computational Linguistics",
    url = "https://aclanthology.org/D18-2012/",
    doi = "10.18653/v1/D18-2012",
    pages = "66--71"
}

@inproceedings{papineni-etal-2002-bleu,
    title = "{B}leu: a Method for Automatic Evaluation of Machine Translation",
    author = "Papineni, Kishore  and
      Roukos, Salim  and
      Ward, Todd  and
      Zhu, Wei-Jing",
    editor = "Isabelle, Pierre  and
      Charniak, Eugene  and
      Lin, Dekang",
    booktitle = "Proceedings of the 40th Annual Meeting of the Association for Computational Linguistics",
    month = jul,
    year = "2002",
    address = "Philadelphia, Pennsylvania, USA",
    publisher = "Association for Computational Linguistics",
    url = "https://aclanthology.org/P02-1040/",
    doi = "10.3115/1073083.1073135",
    pages = "311--318"
}

@inproceedings{popovic-2017-chrf,
    title = "chr{F}++: words helping character n-grams",
    author = "Popovi{\'c}, Maja",
    editor = "Bojar, Ond{\v{r}}ej  and
      Buck, Christian  and
      Chatterjee, Rajen  and
      Federmann, Christian  and
      Graham, Yvette  and
      Haddow, Barry  and
      Huck, Matthias  and
      Yepes, Antonio Jimeno  and
      Koehn, Philipp  and
      Kreutzer, Julia",
    booktitle = "Proceedings of the Second Conference on Machine Translation",
    month = sep,
    year = "2017",
    address = "Copenhagen, Denmark",
    publisher = "Association for Computational Linguistics",
    url = "https://aclanthology.org/W17-4770/",
    doi = "10.18653/v1/W17-4770",
    pages = "612--618"
}

@inproceedings{juraska-etal-2024-metricx,
    title = "{M}etric{X}-24: The {G}oogle Submission to the {WMT} 2024 Metrics Shared Task",
    author = "Juraska, Juraj  and
      Deutsch, Daniel  and
      Finkelstein, Mara  and
      Freitag, Markus",
    editor = "Haddow, Barry  and
      Kocmi, Tom  and
      Koehn, Philipp  and
      Monz, Christof",
    booktitle = "Proceedings of the Ninth Conference on Machine Translation",
    month = nov,
    year = "2024",
    address = "Miami, Florida, USA",
    publisher = "Association for Computational Linguistics",
    url = "https://aclanthology.org/2024.wmt-1.35/",
    doi = "10.18653/v1/2024.wmt-1.35",
    pages = "492--504"
}

@inproceedings{lavie-etal-2025-findings,
    title = "Findings of the {WMT}25 Shared Task on Automated Translation Evaluation Systems: Linguistic Diversity is Challenging and References Still Help",
    author = "Lavie, Alon  and
      Hanneman, Greg  and
      Agrawal, Sweta  and
      Kanojia, Diptesh  and
      Lo, Chi-Kiu  and
      Zouhar, Vil{\'e}m  and
      Blain, Frederic  and
      Zerva, Chrysoula  and
      Avramidis, Eleftherios  and
      Deoghare, Sourabh  and
      Sindhujan, Archchana  and
      Wang, Jiayi  and
      Adelani, David Ifeoluwa  and
      Thompson, Brian  and
      Kocmi, Tom  and
      Freitag, Markus  and
      Deutsch, Daniel",
    editor = "Haddow, Barry  and
      Kocmi, Tom  and
      Koehn, Philipp  and
      Monz, Christof",
    booktitle = "Proceedings of the Tenth Conference on Machine Translation",
    month = nov,
    year = "2025",
    address = "Suzhou, China",
    publisher = "Association for Computational Linguistics",
    url = "https://aclanthology.org/2025.wmt-1.24/",
    doi = "10.18653/v1/2025.wmt-1.24",
    pages = "436--483",
    ISBN = "979-8-89176-341-8"
}

@article{kuhn-2014-survey,
    title = "A Survey and Classification of Controlled Natural Languages",
    author = "Kuhn, Tobias",
    journal = "Computational Linguistics",
    volume = "40",
    number = "1",
    month = mar,
    year = "2014",
    address = "Cambridge, MA",
    publisher = "MIT Press",
    url = "https://aclanthology.org/J14-1005/",
    doi = "10.1162/COLI_a_00168",
    pages = "121--170"
}

@inproceedings{laubli-etal-2018-machine,
    title = "Has Machine Translation Achieved Human Parity? A Case for Document-level Evaluation",
    author = {L{\"a}ubli, Samuel  and
      Sennrich, Rico  and
      Volk, Martin},
    editor = "Riloff, Ellen  and
      Chiang, David  and
      Hockenmaier, Julia  and
      Tsujii, Jun{'}ichi",
    booktitle = "Proceedings of the 2018 Conference on Empirical Methods in Natural Language Processing",
    month = oct # "-" # nov,
    year = "2018",
    address = "Brussels, Belgium",
    publisher = "Association for Computational Linguistics",
    url = "https://aclanthology.org/D18-1512/",
    doi = "10.18653/v1/D18-1512",
    pages = "4791--4796"
}

@article{wandel2015many,
  title={How many people speak Esperanto? Esperanto on the Web},
  author={Wandel, Amri},
  journal={Interdisciplinary Description of Complex Systems: INDECS},
  volume={13},
  number={2},
  pages={318--321},
  year={2015},
  publisher={Hrvatsko interdisciplinarno dru{\v{s}}tvo}
}

@book{eberhard2026ethnologue,
  editor    = {Eberhard, David M. and Simons, Gary F. and Robinson, Alison J.},
  title     = {Ethnologue: Languages of the World},
  edition   = {29},
  year      = {2026},
  address   = {Dallas, Texas},
  publisher = {SIL Global},
  url       = {https://www.ethnologue.com/language/epo/},
  note      = {Online version}
}

@misc{unesco1954resolution,
  author       = {{UNESCO}},
  title        = {Records of the General Conference, Eighth Session, Montevideo 1954; Resolutions},
  year         = {1954},
  howpublished = {UNESDOC Database},
  note         = {p.~36. Archived from the original (PDF) on February 2, 2011. Retrieved May 16, 2018},
  url          = {https://unesdoc.unesco.org/}
}

@article{parkvall2010european,
  title={How European is Esperanto?: A typological study},
  author={Parkvall, Mikael},
  journal={Language Problems and Language Planning},
  volume={34},
  number={1},
  pages={63--79},
  year={2010},
  publisher={John Benjamins}
}

@article{hutchins1992introduction,
  title={An introduction to machine translation},
  author={Hutchins, William John and Somers, Harold L},
  journal={(No Title)},
  year={1992}
}

@article{team2024gemma,
  title={Gemma 2: Improving open language models at a practical size},
  author={Team, Gemma and Riviere, Morgane and Pathak, Shreya and Sessa, Pier Giuseppe and Hardin, Cassidy and Bhupatiraju, Surya and Hussenot, L{\'e}onard and Mesnard, Thomas and Shahriari, Bobak and Ram{\'e}, Alexandre and others},
  journal={arXiv preprint arXiv:2408.00118},
  year={2024}
}

@article{vaswani2017attention,
  title={Attention is all you need},
  author={Vaswani, Ashish and Shazeer, Noam and Parmar, Niki and Uszkoreit, Jakob and Jones, Llion and Gomez, Aidan N and Kaiser, {\L}ukasz and Polosukhin, Illia},
  journal={Advances in neural information processing systems},
  volume={30},
  year={2017}
}

@inproceedings{lui-baldwin-2012-langid,
    title = "langid.py: An Off-the-shelf Language Identification Tool",
    author = "Lui, Marco  and
      Baldwin, Timothy",
    editor = "Zhang, Min",
    booktitle = "Proceedings of the {ACL} 2012 System Demonstrations",
    month = jul,
    year = "2012",
    address = "Jeju Island, Korea",
    publisher = "Association for Computational Linguistics",
    url = "https://aclanthology.org/P12-3005/",
    pages = "25--30"
}

@inproceedings{Aycock24,
title = {Can {{LLMs Really Learn}} to {{Translate}} a {{Low-Resource Language}} from {{One Grammar Book}}?},
booktitle = {Proceedings of {{The Thirteenth International Conference}} on {{Learning Representations}}},
author = {Aycock, Seth and Stap, David and Wu, Di and Monz, Christof and Sima'an, Khalil},
year = 2025,
month = apr,
pages = {12334--12357},
adress = {Singapore},
url = {https://proceedings.iclr.cc/paper_files/paper/2025/file/20f44da80080d76bbc35bca0027f14e6-Paper-Conference.pdf},
urldate = {2025-02-20}
}

@inproceedings{deutsch-etal-2023-ties,
    title = "Ties Matter: Meta-Evaluating Modern Metrics with Pairwise Accuracy and Tie Calibration",
    author = "Deutsch, Daniel  and
      Foster, George  and
      Freitag, Markus",
    editor = "Bouamor, Houda  and
      Pino, Juan  and
      Bali, Kalika",
    booktitle = "Proceedings of the 2023 Conference on Empirical Methods in Natural Language Processing",
    month = dec,
    year = "2023",
    address = "Singapore",
    publisher = "Association for Computational Linguistics",
    url = "https://aclanthology.org/2023.emnlp-main.798/",
    doi = "10.18653/v1/2023.emnlp-main.798",
    pages = "12914--12929"
}

@inproceedings{de-gibert-etal-2025-scaling,
    title = "Scaling Low-Resource {MT} via Synthetic Data Generation with {LLM}s",
    author = {de Gibert, Ona  and
      Attieh, Joseph  and
      Vahtola, Teemu  and
      Aulamo, Mikko  and
      Li, Zihao  and
      V{\'a}zquez, Ra{\'u}l  and
      Hu, Tiancheng  and
      Tiedemann, J{\"o}rg},
    editor = "Christodoulopoulos, Christos  and
      Chakraborty, Tanmoy  and
      Rose, Carolyn  and
      Peng, Violet",
    booktitle = "Proceedings of the 2025 Conference on Empirical Methods in Natural Language Processing",
    month = nov,
    year = "2025",
    address = "Suzhou, China",
    publisher = "Association for Computational Linguistics",
    url = "https://aclanthology.org/2025.emnlp-main.1408/",
    doi = "10.18653/v1/2025.emnlp-main.1408",
    pages = "27674--27692",
    ISBN = "979-8-89176-332-6"
}

@inproceedings{tapo-etal-2025-bayelemabaga,
    title = "Bayelemabaga: Creating Resources for {B}ambara {NLP}",
    author = "Tapo, Allahsera Auguste  and
      Assogba, Kevin  and
      Homan, Christopher M  and
      Rafique, M. Mustafa  and
      Zampieri, Marcos",
    editor = "Chiruzzo, Luis  and
      Ritter, Alan  and
      Wang, Lu",
    booktitle = "Proceedings of the 2025 Conference of the Nations of the Americas Chapter of the Association for Computational Linguistics: Human Language Technologies (Volume 1: Long Papers)",
    month = apr,
    year = "2025",
    address = "Albuquerque, New Mexico",
    publisher = "Association for Computational Linguistics",
    url = "https://aclanthology.org/2025.naacl-long.602/",
    doi = "10.18653/v1/2025.naacl-long.602",
    pages = "12060--12070",
    ISBN = "979-8-89176-189-6"
}

@inproceedings{scalvini-etal-2025-rethinking,
    title = "Rethinking Low-Resource {MT:} The Surprising Effectiveness of Fine-Tuned Multilingual Models in the {LLM} Age",
    author = "Scalvini, Barbara  and
      Debess, Iben Nyholm  and
      Simonsen, Annika  and
      Einarsson, Hafsteinn",
    editor = "Johansson, Richard  and
      Stymne, Sara",
    booktitle = "Proceedings of the Joint 25th Nordic Conference on Computational Linguistics and 11th Baltic Conference on Human Language Technologies (NoDaLiDa/Baltic-HLT 2025)",
    month = mar,
    year = "2025",
    address = "Tallinn, Estonia",
    publisher = "University of Tartu Library",
    url = "https://aclanthology.org/2025.nodalida-1.62/",
    pages = "609--621",
    ISBN = "978-9908-53-109-0"
}

@inproceedings{kocmi2025findings,
  title={Findings of the wmt25 general machine translation shared task: Time to stop evaluating on easy test sets},
  author={Kocmi, Tom and Artemova, Ekaterina and Avramidis, Eleftherios and Bawden, Rachel and Bojar, Ond{\v{r}}ej and Dranch, Konstantin and Dvorkovich, Anton and Dukanov, Sergey and Fishel, Mark and Freitag, Markus and others},
  booktitle={Proceedings of the Tenth Conference on Machine Translation},
  pages={355--413},
  year={2025}
}

@inproceedings{xue-etal-2021-mt5,
    title = "m{T}5: A Massively Multilingual Pre-trained Text-to-Text Transformer",
    author = "Xue, Linting  and
      Constant, Noah  and
      Roberts, Adam  and
      Kale, Mihir  and
      Al-Rfou, Rami  and
      Siddhant, Aditya  and
      Barua, Aditya  and
      Raffel, Colin",
    editor = "Toutanova, Kristina  and
      Rumshisky, Anna  and
      Zettlemoyer, Luke  and
      Hakkani-Tur, Dilek  and
      Beltagy, Iz  and
      Bethard, Steven  and
      Cotterell, Ryan  and
      Chakraborty, Tanmoy  and
      Zhou, Yichao",
    booktitle = "Proceedings of the 2021 Conference of the North American Chapter of the Association for Computational Linguistics: Human Language Technologies",
    month = jun,
    year = "2021",
    address = "Online",
    publisher = "Association for Computational Linguistics",
    url = "https://aclanthology.org/2021.naacl-main.41/",
    doi = "10.18653/v1/2021.naacl-main.41",
    pages = "483--498"
}

@inproceedings{conneau2020unsupervised,
  title={Unsupervised cross-lingual representation learning at scale},
  author={Conneau, Alexis and Khandelwal, Kartikay and Goyal, Naman and Chaudhary, Vishrav and Wenzek, Guillaume and Guzm{\'a}n, Francisco and Grave, Edouard and Ott, Myle and Zettlemoyer, Luke and Stoyanov, Veselin},
  booktitle={Proceedings of the 58th annual meeting of the association for computational linguistics},
  pages={8440--8451},
  year={2020}
}

@inproceedings{de-gibert-etal-2024-new,
    title = "A New Massive Multilingual Dataset for High-Performance Language Technologies",
    author = {de Gibert, Ona  and
      Nail, Graeme  and
      Arefyev, Nikolay  and
      Ba{\~n}{\'o}n, Marta  and
      van der Linde, Jelmer  and
      Ji, Shaoxiong  and
      Zaragoza-Bernabeu, Jaume  and
      Aulamo, Mikko  and
      Ram{\'i}rez-S{\'a}nchez, Gema  and
      Kutuzov, Andrey  and
      Pyysalo, Sampo  and
      Oepen, Stephan  and
      Tiedemann, J{\"o}rg},
    editor = "Calzolari, Nicoletta  and
      Kan, Min-Yen  and
      Hoste, Veronique  and
      Lenci, Alessandro  and
      Sakti, Sakriani  and
      Xue, Nianwen",
    booktitle = "Proceedings of the 2024 Joint International Conference on Computational Linguistics, Language Resources and Evaluation (LREC-COLING 2024)",
    month = may,
    year = "2024",
    address = "Torino, Italia",
    publisher = "ELRA and ICCL",
    url = "https://aclanthology.org/2024.lrec-main.100/",
    pages = "1116--1128"
}

@inproceedings{burchell-etal-2025-expanded,
    title = "An Expanded Massive Multilingual Dataset for High-Performance Language Technologies ({HPLT})",
    author = {Burchell, Laurie  and
      de Gibert, Ona  and
      Arefyev, Nikolay  and
      Aulamo, Mikko  and
      Ba{\~n}{\'o}n, Marta  and
      Chen, Pinzhen  and
      Fedorova, Mariia  and
      Guillou, Liane  and
      Haddow, Barry  and
      Haji{\v{c}}, Jan  and
      Helcl, Jind{\v{r}}ich  and
      Henriksson, Erik  and
      Klimaszewski, Mateusz  and
      Komulainen, Ville  and
      Kutuzov, Andrey  and
      Kyt{\"o}niemi, Joona  and
      Laippala, Veronika  and
      M{\ae}hlum, Petter  and
      Malik, Bhavitvya  and
      Mehryary, Farrokh  and
      Mikhailov, Vladislav  and
      Moghe, Nikita  and
      Myntti, Amanda  and
      O{'}Brien, Dayy{\'a}n  and
      Oepen, Stephan  and
      Pal, Proyag  and
      Piha, Jousia  and
      Pyysalo, Sampo  and
      Ram{\'i}rez-S{\'a}nchez, Gema  and
      Samuel, David  and
      Stepachev, Pavel  and
      Tiedemann, J{\"o}rg  and
      Vari{\v{s}}, Du{\v{s}}an  and
      Vojt{\v{e}}chov{\'a}, Tereza  and
      Zaragoza-Bernabeu, Jaume},
    editor = "Che, Wanxiang  and
      Nabende, Joyce  and
      Shutova, Ekaterina  and
      Pilehvar, Mohammad Taher",
    booktitle = "Proceedings of the 63rd Annual Meeting of the Association for Computational Linguistics (Volume 1: Long Papers)",
    month = jul,
    year = "2025",
    address = "Vienna, Austria",
    publisher = "Association for Computational Linguistics",
    url = "https://aclanthology.org/2025.acl-long.854/",
    doi = "10.18653/v1/2025.acl-long.854",
    pages = "17452--17485",
    ISBN = "979-8-89176-251-0"
}

@article{oepen2025hplt,
  title={HPLT 3.0: Very Large-Scale Multilingual Resources for LLM and MT. Mono-and Bi-lingual Data, Multilingual Evaluation, and Pre-Trained Models},
  author={Oepen, Stephan and Arefev, Nikolay and Aulamo, Mikko and Ba{\~n}{\'o}n, Marta and Buljan, Maja and Burchell, Laurie and Charpentier, Lucas and Chen, Pinzhen and Fedorova, Mariya and de Gibert, Ona and others},
  journal={arXiv preprint arXiv:2511.01066},
  year={2025}
}

@article{kudugunta2023madlad,
  title={Madlad-400: A multilingual and document-level large audited dataset},
  author={Kudugunta, Sneha and Caswell, Isaac and Zhang, Biao and Garcia, Xavier and Xin, Derrick and Kusupati, Aditya and Stella, Romi and Bapna, Ankur and Firat, Orhan},
  journal={Advances in Neural Information Processing Systems},
  volume={36},
  pages={67284--67296},
  year={2023}
}

\appendix

\section{Training details} \label{appen:hyperparam}
Below, we list the hyperparameters used during training for Transformer-base (Table~\ref{tab:tf-base}), Transformer-tiny (Table~\ref{tab:tf-tiny}), and Llama fine-tuning (Table~\ref{tab:llama}).
We also report the architectures for the tiny and base Transformer models in Table~\ref{tab:architectures}.

\begin{table}[ht!]
    \centering
    \adjustbox{max width=\linewidth}{
    \begin{tabular}{lrr}
    \toprule
    & base & tiny  \\
     \midrule
     \textit{N}$_{enc}$ & 6 & 6 \\
    \textit{N}$_{dec}$ & 6  & 2 \\
     \textit{d}$_{emb}$ & 1024 & 256 \\
     \textit{d}$_{ff}$&2048 & 1536 \\
     \textit{h} &8& 8\\
    \midrule
     Params (M) &  60.6  & 17.4 \\
     Size (MB) & 232 & 67 \\
     \bottomrule
    \end{tabular}}  
    \caption{Transformer architectures for base and tiny. The table lists the number of encoder and decoder layers (\textit{N}$_{enc}$ and \textit{N}$_{dec}$), embedding dimensions (\textit{d}$_{emb}$), feed-forward dimensions (\textit{d}$_{ff}$), number of attention heads (\textit{h}), parameters in millions, and model size in MB.}
    \label{tab:architectures}
\end{table}

\newpage

\begin{table}[h!]
\centering
\begin{tabular}{ll}
\toprule
Hyperparameter & Value\\
\midrule
Optimizer & Adam \\
Adam $\beta$ & (0.9, 0.98) \\
Adam $\epsilon$ & 1e-9 \\
Learning rate & 3e-4 \\
LR warmup & 16000 steps \\
LR decay strategy & epoch+stalled \\
LR decay start & (10 epochs, 1 stalled) \\
Optimizer delay & 1  \\
\addlinespace
Validation frequency & 2500 updates \\
Early stopping & 10 (on perplexity) \\
\addlinespace
Seed & 1111 \\
\bottomrule
\end{tabular}
\caption{Hyperparameters for Transformer-base}
\label{tab:tf-base}
\end{table}

\begin{table}[h!]
\centering
\begin{tabular}{ll}
\toprule
Hyperparameter & Value\\
\midrule
Optimizer & Adam \\
Adam $\beta$ & (0.9, 0.98) \\
Adam $\epsilon$ & 1e-9 \\
Learning rate & 3e-4 \\
LR warmup & 16000 steps \\
LR decay strategy & epoch+stalled \\
LR decay start & (10 epochs, 1 stalled) \\
Optimizer delay & 2 \\
LR inv-sqrt factor & 32000 \\
\addlinespace
Validation frequency & 5000 updates \\
Early stopping & 20 (on cross-entropy) \\
\addlinespace
Seed & 0 \\
\bottomrule
\end{tabular}
\caption{Hyperparameters for Transfomer-tiny}
\label{tab:tf-tiny}
\end{table}

\begin{table}[h!]
\centering
\begin{tabular}{ll}
\toprule
Hyperparameter & Value \\
\midrule
Learning Rate & 5e-05 \\
LR Scheduler Type & Linear \\
Warmup Ratio & 0.3 \\
Weight Decay & 0.0 \\
\addlinespace
Per Device Train Batch Size & 4 \\
Gradient Accumulation Steps & 4 \\
Number of Train Epochs & 1 \\
\addlinespace
LoRA Rank & 16 \\
LoRA Alpha & 32 \\
\addlinespace
Seed & 123 \\
\bottomrule
\end{tabular}
\caption{Hyperparameters for Llama fine-tuning}
\label{tab:llama}
\end{table}

\begin{table}[ht!]
\centering
\begin{tabular}{ll}
\toprule
Hyperparameter & Value \\
\midrule
Optimizer & Adafactor \\
Learning Rate & 1e-4 \\
LR Scheduler Type & Constant \\
LR Warmup & 1000 steps \\
Weight Decay & 1e-3 \\
\addlinespace
Per Device Train Batch Size & 32 \\
Gradient Accumulation Steps & 2 \\
Maximum Sequence Length & 128 \\
Number of Train Epochs & 4 \\
\addlinespace
Validation Frequency & 1000 updates \\
Early Stopping & 5 \\
\addlinespace
Seed & 42 \\
\bottomrule
\end{tabular}
\caption{Hyperparameters for NLLB fine-tuning}
\label{tab:nllb-ft}
\end{table}

\section{Details of the NLLB fine-tuning experiments}
\label{appen:nllb}
\textcolor{black}{We conducted exploratory fine-tuning experiments with Hugging Face implementations of NLLB. We fine-tuned the 600M- and 3.3B-parameter models in a multilingual setup, training one model for each Esperanto language direction. The training data were the same as those used for Marian. Details of the fine-tuning hyperparameters are given in Table~\ref{tab:nllb-ft}. Under this configuration, fine-tuning did not improve over the base NLLB models, with results generally comparable to or slightly below those of the original checkpoints.}

\section{Automatic Evaluation Results}
\label{appen:metrics}
We report automatic evaluation results using BLEU (Table~\ref{tab:bleu}), COMET (Table~\ref{tab:comet}), and MetricX (Table~\ref{tab:metricx}).

\begin{table*}[h]
    \centering
    \adjustbox{width=0.85\linewidth}{
    \begin{tabular}{lccccccc}
\toprule
 & eo-en & eo-es & eo-ca & en-eo & es-eo & ca-eo  \\
 \midrule
\rowcolor{gray!15} Rule-based MT & & & & & & \\
\hspace{5mm}  Apertium & \cellcolor{myred!45}19.94 & - & - & 20.80 & 10.99 & 14.34 \\
\rowcolor{gray!15}Neural MT & & && & &  \\
\hspace{5mm}  NLLB-200-distilled-600M & 43.04 & 21.74 & 27.98 & 31.86 & 18.18 & 24.20 \\
\hspace{5mm}  NLLB-200-1.3B & 44.66 & 23.37 & 30.82 & 33.11 & 18.83 & 24.26 \\
\hspace{5mm}  NLLB-200-distilled-1.3B & 45.49 & \cellcolor{mygreen!45}23.63 & 31.63 & \cellcolor{mygreen!45}33.52 & 18.98 & 24.26 \\
\hspace{5mm}  NLLB-200-3.3B & \cellcolor{mygreen!45}46.05 & 23.50 & \cellcolor{mygreen!45}32.10 & 33.47 & \cellcolor{mygreen!45}19.25 & \cellcolor{mygreen!45}24.54 \\
\rowcolor{gray!15}General-purpose LLMs & && & &  & \\
\hspace{5mm}  Llama-3.1-8B-Instruct & 40.05 & 19.08 & 23.03 & 27.57 & 13.52 & 19.70 \\
\rowcolor{gray!15}MT-tuned LLMs & & && & &  \\
\hspace{5mm}  TowerInstruct-7B-v0.2 & 27.28 & \cellcolor{myred!45}14.61 & \cellcolor{myred!45}0.35 & \cellcolor{myred!45}4.66 & \cellcolor{myred!45}3.08 & \cellcolor{myred!45}2.72 \\
\hspace{5mm} Tower-Plus-9B & 42.74 & 21.43 & 23.01 & 17.80 & 10.80 & 14.10 \\
\midrule
\vspace{5mm} \\
\toprule
 & eo-en & eo-es & eo-ca & en-eo & es-eo & ca-eo  \\
 \midrule
\rowcolor{gray!15}Neural MT from Scratch & & & & & & \\
\hspace{5mm} Transformer-base (60.6M) & 37.47 & \cellcolor{mygreen!45}{20.00} & \cellcolor{mygreen!45}{28.35} & \cellcolor{mygreen!45}{26.42} & 16.25 & 21.43 \\
\hspace{5mm} Transformer-tiny (17.4M) & \cellcolor{myred!45}{33.13} & \cellcolor{myred!45}{18.49} & \cellcolor{myred!45}{23.58} & 25.69 & \cellcolor{myred!45}{15.04} & \cellcolor{myred!45}{20.78} \\
\rowcolor{gray!15}Fine-tuned General-purpose LLMs & & & & & & \\
\hspace{5mm} Llama-3.1-8B-Instruct-FT & \cellcolor{mygreen!45}{38.55} & 19.61 & 24.98 & \cellcolor{myred!45}{25.14} & \cellcolor{mygreen!45}{17.17} & \cellcolor{mygreen!45}{22.33} \\
\bottomrule
    \end{tabular}
    }
    \caption{BLEU scores for our benchmarked (above) and trained models (below).  \colorbox{mygreen!45}{Best} and
\colorbox{myred!45}{worst} scores are highlighted for each language direction.}
    \label{tab:bleu}
\end{table*}

\begin{table*}[h]
    \centering
    \adjustbox{width=0.85\linewidth}{
    \begin{tabular}{lccccccc}
\toprule
 & eo-en & eo-es & eo-ca & en-eo & es-eo & ca-eo  \\
 \midrule
\rowcolor{gray!15} Rule-based MT & & & & & & \\
\hspace{5mm} Apertium & \cellcolor{myred!45}70.43 & - & - & 77.67 & 76.02 & 71.77 \\
\rowcolor{gray!15}Neural MT & & && & &  \\
\hspace{5mm}  NLLB-200-distilled-600M & 87.80 & 82.10 & 81.99 & 88.92 & 86.09 & 85.89 \\
\hspace{5mm}  NLLB-200-1.3B & 88.58 & 83.71 & 84.22 & 89.74 & 86.80 & 86.62 \\
\hspace{5mm}  NLLB-200-distilled-1.3B & 88.72 & 83.85 & 84.52 & \cellcolor{mygreen!45}89.85 & 86.93 & 86.24 \\
\hspace{5mm}  NLLB-200-3.3B & \cellcolor{mygreen!45}88.82 & \cellcolor{mygreen!45}84.03 & \cellcolor{mygreen!45}85.07 & 89.82 & \cellcolor{mygreen!45}86.96 & \cellcolor{mygreen!45}86.80 \\
\rowcolor{gray!15}General-purpose LLMs & && & &  & \\
\hspace{5mm}  Llama-3.1-8B-Instruct & 87.23 & 80.69 & 77.93 & 87.09 & 82.64 & 82.92 \\
\rowcolor{gray!15}MT-tuned LLMs & & && & &  \\
\hspace{5mm}  TowerInstruct-7B-v0.2 & 75.54 & \cellcolor{myred!45}65.93 & \cellcolor{myred!45}34.61 & \cellcolor{myred!45}50.86 & \cellcolor{myred!45}54.82 & \cellcolor{myred!45}56.36 \\
\hspace{5mm}  Tower-Plus-9B & 87.19 & 81.79 & 77.79 & 76.62 & 73.71 & 73.63 \\
\midrule
\vspace{5mm} \\
\toprule
 & eo-en & eo-es & eo-ca & en-eo & es-eo & ca-eo  \\
 \midrule
\rowcolor{gray!15}Neural MT from Scratch & & & & & & \\
\hspace{5mm}  Transformer-base (60.6M) & 86.07 & 80.66 & \cellcolor{mygreen!45}82.30 &  \cellcolor{mygreen!45}85.95 & 83.51 & 80.99 \\
\hspace{5mm}  Transformer-tiny (17.4M) &  \cellcolor{myred!45}81.94 &  \cellcolor{myred!45}76.44 &  \cellcolor{myred!45}73.43 &  \cellcolor{myred!45}84.12 &  \cellcolor{myred!45}81.59 &  \cellcolor{myred!45}80.72 \\
\rowcolor{gray!15}Fine-tuned General-purpose LLMs & & & & & & \\
\hspace{5mm}  Llama-3.1-8B-Instruct-FT &  \cellcolor{mygreen!45}86.97 &  \cellcolor{mygreen!45}81.27 & 81.65 & 85.61 &  \cellcolor{mygreen!45}85.40 &  \cellcolor{mygreen!45}85.68 \\
\bottomrule
    \end{tabular}
    }
    \caption{COMET scores for our benchmarked (above) and trained models (below).  \colorbox{mygreen!45}{Best} and
\colorbox{myred!45}{worst} scores are highlighted for each language direction.}
    \label{tab:comet}
\end{table*}

\begin{table*}[h]
    \centering
    \adjustbox{width=0.85\linewidth}{
    \begin{tabular}{lccccccc}
\toprule
 & eo-en & eo-es & eo-ca & en-eo & es-eo & ca-eo  \\
 \midrule
\rowcolor{gray!15} Rule-based MT & & & & & & \\
\hspace{5mm}  Apertium & \cellcolor{myred!45}9.90 & - & - & 8.04 & 7.36 & 8.68\\
\rowcolor{gray!15}Neural MT & & && & &  \\
\hspace{5mm}  NLLB-200-distilled-600M & 2.88 & 3.26 & 4.15 & 4.09 & 4.56 & 4.90\\
\hspace{5mm}  NLLB-200-1.3B & 2.57 & 2.73 & 3.36 & 3.74 & 4.19 & 4.71\\
\hspace{5mm}  NLLB-200-distilled-1.3B & 2.57 & 2.67 & 3.35 & \cellcolor{mygreen!45}3.59 & \cellcolor{mygreen!45}4.08 & 4.85 \\
\hspace{5mm}  NLLB-200-3.3B & \cellcolor{mygreen!45}2.52 & \cellcolor{mygreen!45}2.63 & \cellcolor{mygreen!45}3.12 & 3.64 & 4.10 & \cellcolor{mygreen!45}4.57 \\
\rowcolor{gray!15}General-purpose LLMs & && & &  & \\
\hspace{5mm}  Llama-3.1-8B-Instruct & 3.05 & 3.62 & 5.12 & 4.98 & 5.67 & 5.87\\
\rowcolor{gray!15}MT-tuned LLMs & & && & &  \\
\hspace{5mm}  TowerInstruct-7B-v0.2 & 7.33 & \cellcolor{myred!45}8.81 & \cellcolor{myred!45}13.03 & \cellcolor{myred!45}12.48 & \cellcolor{myred!45}14.18 & \cellcolor{myred!45}12.37 \\
\hspace{5mm}  Tower-Plus-9B & 3.08 & 3.28 & 5.10 & 7.82 & 8.36 & 8.54 \\
\midrule
\vspace{5mm} \\
\toprule
 & eo-en & eo-es & eo-ca & en-eo & es-eo & ca-eo  \\
 \midrule
\rowcolor{gray!15}Neural MT from Scratch & & & & & & \\
\hspace{5mm}  Transformer-base (60.6M) & 3.47 &  \cellcolor{mygreen!45}3.61 &  \cellcolor{mygreen!45}3.94 &  \cellcolor{mygreen!45}4.68 & 5.34 & 6.73\\
\hspace{5mm}  Transformer-tiny (17.4M) & \cellcolor{myred!45}5.15 & \cellcolor{myred!45}5.11 & \cellcolor{myred!45}6.89 & \cellcolor{myred!45}5.53 & \cellcolor{myred!45}6.03 & \cellcolor{myred!45}6.73\\
\rowcolor{gray!15}Fine-tuned General-purpose LLMs & & & & & & \\
\hspace{5mm}  Llama-3.1-8B-Instruct-FT &  \cellcolor{mygreen!45}3.33 & 3.77 & 4.42 & 5.50 &  \cellcolor{mygreen!45}4.86 &  \cellcolor{mygreen!45}5.15\\
\bottomrule
    \end{tabular}
    }
    \caption{MetricX scores for our benchmarked (above) and trained models (below).  \colorbox{mygreen!45}{Best} and
\colorbox{myred!45}{worst} scores are highlighted for each language direction. Lowest is best for this metric.}
    \label{tab:metricx}
\end{table*}

\newpage

\section{Annotation Guidelines}
\label{appen:guidelines}

Figure \ref{fig:guidelines} shows the guidelines presented to the annotators for the human evaluation task.

\begin{figure}[h!]
    \centering
    \fbox{%
    \begin{minipage}{0.9\linewidth}
        \small
        \textbf{Annotation Task.} For each source sentence, you will see three possible translations (T1, T2, and T3). Read them carefully and indicate which translation is the best and which is the worst. Write 1, 2, or 3 in the corresponding columns.

        \vspace{2mm}
        You may also add a short optional comment explaining your decision (e.g., if something sounds unnatural or contains a clear error). There is no need for technical analysis, simply choose the translation that sounds most natural and most faithful to the original meaning.
    \end{minipage}%
    }
    \caption{Annotation guidelines shown to the human annotator.}
    \label{fig:guidelines}
\end{figure}

\section{Qualitative Error Analysis}
\label{appen:error}
Tables~\ref{tab:errors-eo-es} and \ref{tab:errors-es-eo} present illustrative examples of recurring errors in the evaluated models for translation into Spanish and Esperanto, respectively.

\begin{table*}[]
    \centering
    \small
    \begin{tabular}{lp{0.4\linewidth}p{0.4\linewidth}}
    \toprule
    Error Type & Source & Translation \\
    \midrule
    \rowcolor{gray!15}\multicolumn{3}{c}{Transformer-base}\\
    Lexical & \textit{La epidemio igis la baratan registaron entrepreni tiajn rimedojn, kiel instalado de porkokaptiloj en la serioze damaĝitaj areoj, disdonado de miloj da \colorbox{mygreen!45}{kontraŭmoskitaj} retoj kaj ŝprucado de pesticidoj.} & \textit{La epidemia llevó al gobierno indio a emprender tales medios, como la instalación de trampas de cerdo en las áreas seriamente dañadas, la distribución de miles de redes \colorbox{myred!45}{anti-moscoquios} y el brote de pesticidas.} \\
    Grammatical & \textit{Malgraŭ foresto de cunama minaco, loĝantoj \colorbox{mygreen!45}{ekpanikis} kaj komencis forlasadi siajn komercojn kaj hejmojn.} & \textit{A pesar de la ausencia de la amenaza tsunami, los residentes \colorbox{myred!45}{comenzaron a pánico} y comenzaron a dejar sus negocios y casas.}\\ 
    Untranslated & \textit{\colorbox{mygreen!45}{Gregon} konsistigas de unu ĝis tri parencaj plenkreskaj maskloj kaj multaj ĝis tridek inoj kaj idoj.} & \textit{\colorbox{myred!45}{Gregon} consiste de uno a tres varones adultos relacionados y muchos a treinta hembras y crías.} \\
    \rowcolor{gray!15}\multicolumn{3}{c}{NLLB-200-3.3B}\\
     Mistranslation & \textit{\colorbox{mygreen!45}{Nekredantoj} vivantaj en kristanaj landoj aŭ aliloke en la tuta mondo adoptis multajn el la tradicioj pri ĉi tiu festo.} & \textit{Los \colorbox{myred!45}{incrédulos} que viven en países cristianos o en otras partes del mundo han adoptado muchas de las tradiciones de esta fiesta.} \\
    & \textit{Malsupreniraj vintrosportoj, kiuj inkluzivas skiadon kaj neĝtabuladon, estas popularaj sportoj, en kiuj oni glitas malsupren sur neĝokovrita tereno havante skiojn aŭ \colorbox{mygreen!45}{neĝtabulon} fiksitaj al la piedoj.} & \textit{Los deportes de invierno de descenso, que incluyen el esquí y el snowboard, son deportes populares en los que se desliza hacia abajo en un terreno cubierto de nieve con esquís o una \colorbox{myred!45}{tabla de nieve} fijada a los pies.} \\
    & \textit{Pro internaciaj sankcioj, novaj \colorbox{mygreen!45}{flugaparatoj} ne estas aĉeteblaj.} & \textit{Debido a las sanciones internacionales, no se pueden comprar nuevos \colorbox{myred!45}{aparatos de vuelo.}} \\
        \rowcolor{gray!15}\multicolumn{3}{c}{Llama-3.1-8B-Instruct-FT}\\
             Addition & \textit{Ni penos malaltigi eligon de karbona dioksido po unuo de MEP je rimarkinda diferenco ĝis la jaro 2020, kompare al la nivelo de 2005", Hu diris.} & \textit{"Haremos todo lo posible para reducir las emisiones de dióxido de carbono por unidad de PIB \colorbox{myred!45}{en un 20\%} hasta el 2020, en comparación con el nivel de 2005", dijo Hu.} \\
             Grammatical & \textit{Post kiam oficialuloj konfirmas identecon de la voĉdonanto, la voĉdonanto enmetas la koverton en la \colorbox{mygreen!45}{balotujon} kaj subskribas la voĉdonan liston.} & \textit{Después de que los funcionarios verifican la identidad del votante, el votante coloca el sobre en \colorbox{myred!45}{el urna} y firma la lista de votación.} \\
             Hallucination & \textit{Gregon konsistigas de unu ĝis \colorbox{mygreen!45}{tri parencaj plenkreskaj maskloj} kaj multaj ĝis tridek inoj kaj idoj.} & \textit{Una colonia consta de \colorbox{myred!45}{un macho adulto} y varias hembras y hasta treinta crías.}\\
             Lexical & \textit{Del Potro havis fruan avantaĝon en la dua serio, sed ankaŭ tiu \colorbox{mygreen!45}{bezonis decidludon}, kiam la rezulto atingis 6-6.} & \textit{Del Potro tenía ventaja en el segundo set, pero necesitó un \colorbox{myred!45}{tiebreaker}, ya que el resultado se igualó en 6-6.} \\
             Mistranslation & \textit{Tio nomiĝas pH de kemiaĵo. Vi povas fari indikanton per uzo de suko de \colorbox{mygreen!45}{ruĝa beto}.} & \textit{Se llama pH de un compuesto. Puedes hacer un indicador con jugo de \colorbox{myred!45}{betún.}} \\
            Omission & \textit{Tigro estas en la sama grupo (Genus Panthera) kun leonoj, leopardoj kaj jaguaroj. \colorbox{myred!45}{Tiuj ĉi kvar katoj estas la solaj, kiuj povas rori.}} & \textit{El tigre pertenece al mismo género (Panthera) que los leones, leopardos y jaguar.} \\

    \bottomrule
    \end{tabular}
    \caption{Illustrative examples of error categories identified in the qualitative analysis for Esperanto into Spanish. Highlighted spans mark erroneous content.}
    \label{tab:errors-eo-es}
\end{table*}

\begin{table*}[]
    \centering
    \small
    \begin{tabular}{lp{0.4\linewidth}p{0.4\linewidth}}
    \toprule
    Error Type & Source & Translation \\
    \midrule
    \rowcolor{gray!15}\multicolumn{3}{c}{Transformer-base}\\
     & \\ 
    Grammatical & \textit{En verdad, el formato de 35 mm es \colorbox{mygreen!45}{algo confuso}, ya que sus medidas son 36 mm de anchura por 24 mm de alto.} & \textit{Fakte, la 35mm formato estas \colorbox{myred!45}{iom konfuzita}, ĉar ĝiaj mezuroj estas 36 mm larĝa de 24 mm alta.}\\
    &  \\
    Mistranslation & \textit{Conforme surge de las imágenes infrarrojas, las \colorbox{mygreen!45}{variaciones de temperatura} durante la noche y el día indican que, probablemente, se trate de cuevas.} & \textit{Ĉar ĝi ekestiĝas de infraruĝaj bildoj, \colorbox{myred!45}{temperaturkatastrofoj} dum la nokto kaj tago indikas ke ili estas verŝajne kavernoj.}\\
    Named Entities & \textit{El día de hoy se originó en el Océano Atlántico la tormenta número diez de su temporada de huracanes en ser nombrada, a la que se denominó \colorbox{mygreen!45}{Tormenta subtropical Jerry.}} & \textit{La atlantika ŝtormo numero dek el sia uragansezono estis nomita \colorbox{myred!45}{Jerry Storm.}} \\
    & \textit{La cueva en sí misma, que perduró en el tiempo, ilustra de forma \colorbox{mygreen!45}{muy realista} las inclinaciones espirituales de Mahoma.} & \textit{La kaverno mem, kiu daŭris en tempo, \colorbox{myred!45}{tre racie} ilustras la spiritajn inklinojn de Mohamedo.} \\
    \rowcolor{gray!15}\multicolumn{3}{c}{NLLB-200-3.3B}\\
   Mistranslation & \textit{Los científicos creen que los ocelotes rastrean y encuentran animales para comer \colorbox{mygreen!45}{(presas)} a través del olfato, oliendo el suelo donde han pasado.} & Sciencistoj kredas, ke la ocelotoj spuras kaj trovas bestojn por manĝi \colorbox{myred!45}{(ofero)} per la flaro, flarante la grundon, kiun ili pasis. \\
    Omission & \textit{A pesar de que había tres personas en el interior de la vivienda \colorbox{myred!45}{en el momento del impacto del automóvil}, ninguna de ellas sufrió lesiones.} & \textit{Kvankam tri homoj estis en la domo, neniu el ili vundiĝis.}  \\
        \rowcolor{gray!15}\multicolumn{3}{c}{Llama-3.1-8B-Instruct-FT}\\
             Grammatical & \textit{Duvall, que está casado y tiene dos hijos adultos, no causó una buena impresión a Miller, \colorbox{mygreen!45}{que fue a quien le relató la historia.}} & \textit{Duvall, kiu estis edziĝinta kaj havis du plenkreskajn infanojn, ne faris bonan impreson al Miller, \colorbox{myred!45}{kiu rakontis la historion.}} \\ 
             Lexical & \textit{El mismo mes, otro avión de pasajeros sobrepasó una pista \colorbox{mygreen!45}{en Mashhad} y colisionó contra una pared, dejando 17 muertos.} & \textit{En la sama monato, alia pasaĝera aviadilo preterflugis \colorbox{myred!45}{Mashhad's} runvojo kaj koliziis kun muro, mortigante 17 homojn.} \\
             Mistranslation & \textit{Serían las \colorbox{mygreen!45}{fuerzas de marea} entre la materia oscura presente en la galaxia y la de la \colorbox{mygreen!45}{Vía Láctea} lo que provocaría el efecto esperado por el grupo de científicos.} & \textit{La \colorbox{myred!45}{fortoj de la maroj} inter la malluma materio en la galaksio kaj tiu de la \colorbox{myred!45}{Vida Laktone} estas tiuj kiuj kaŭzus la atendatan efikon, laŭ la sciencistoj.} \\
            Omission & \textit{Hacia 1976 ya se había restaurado el treinta por ciento \colorbox{myred!45}{del Machu Picchu} y los trabajos continúan en la actualidad.} & \textit{En 1976 estis jam restarigita la triono de la konstrua\^jo kaj la laboroj daŭras.} \\
    \bottomrule
    \end{tabular}
    \caption{Illustrative examples of error categories identified in the qualitative analysis for Spanish into Esperanto. Highlighted spans mark erroneous content.}
    \label{tab:errors-es-eo}
\end{table*}
\end{document}